\documentclass[10pt,twocolumn,letterpaper]{article}

\usepackage{cvpr}              % To produce the CAMERA-READY version
\definecolor{cvprblue}{rgb}{0.21,0.49,0.74}
\usepackage[pagebackref,breaklinks,colorlinks,allcolors=cvprblue]{hyperref}

\usepackage{bm}
\usepackage{caption}
\usepackage{comment}
\usepackage{subcaption}
\usepackage{booktabs}
\usepackage{bold-extra}
\usepackage{wrapfig}
\usepackage{xspace}
\usepackage{dsfont}
\usepackage{colortbl}
\usepackage{graphics}
\usepackage{textcomp}
\usepackage{multirow}
\usepackage{algorithm}
\usepackage{algorithmic}

\usepackage[accsupp]{axessibility}

\definecolor{mygray}{gray}{.92}

\newsavebox{\twosubbox}
\newcommand{\mrtwo}[1]{\multirow{2}{*}{#1}}

\makeatletter
\DeclareRobustCommand\onedot{\futurelet\@let@token\@onedot}
\def\@onedot{\ifx\@let@token.\else.\null\fi\xspace}

\def\eg{\emph{e.g}\onedot}

\def\etc{\emph{etc}\onedot} 
 
\def\etal{\emph{et al}\onedot}
\makeatother

\def\paperID{4567} % *** Enter the Paper ID here
\def\confName{CVPR}
\def\confYear{2025}

\title{DeformCL: Learning Deformable Centerline Representation for Vessel Extraction in 3D Medical Image}

\author{
  \centering % 建议将整体居中
  \begin{tabular}{c} % 使用一个单列的表格来包裹所有内容
    Ziwei Zhao$^{1}$\footnotemark[1] \quad Zhixing Zhang$^{2,5}$\footnotemark[1] \quad Yuhang Liu$^{1}$\footnotemark[2] \quad Zhao Zhang$^{2}$ \\
    Haojun Yu$^{4}$ \quad Dong Wang$^{1}$ \quad Liwei Wang$^{3,4}$ \\
    {\normalsize$^1$Yizhun Medical AI Co., Ltd \quad
    $^2$Center for Data Science, Peking University} \\
    {\normalsize$^3$Center for Machine Learning Research, Peking University} \\
    {\normalsize$^4$State Key Laboratory of General Artificial Intelligence, School of Intelligence Science and Technology, Peking University} \\
    {\normalsize$^5$Pazhou Laboratory (Huangpu), Guangzhou, Guangdong, China} \\
    {\tt\small \{ziwei.zhao, yuhang.liu, dong.wang\}@yizhun-ai.com} \\
    {\tt\small \{zhangzhixing@stu., zhangzh@stu., haojunyu@, wanglw@\}pku.edu.cn}
  \end{tabular}
}

\begin{document}
\maketitle

\renewcommand{\thefootnote}{\fnsymbol{footnote}}
\footnotetext[1]{Equal contribution.}
\footnotetext[2]{Project lead.}

\begin{abstract}
In the field of 3D medical imaging, accurately extracting and representing the blood vessels with curvilinear structures holds paramount importance for clinical diagnosis. Previous methods have commonly relied on discrete representation like mask, often resulting in local fractures or scattered fragments due to the inherent limitations of the per-pixel classification paradigm. In this work, we introduce DeformCL, a new continuous representation based on \textbf{Deform}able \textbf{C}enter\textbf{l}ines, where centerline points act as nodes connected by edges that capture spatial relationships. Compared with previous representations, DeformCL offers three key advantages: natural connectivity, noise robustness, and interaction facility. We present a comprehensive training pipeline structured in a cascaded manner to fully exploit these favorable properties of DeformCL. Extensive experiments on four 3D vessel segmentation datasets demonstrate the effectiveness and superiority of our method. Furthermore, the visualization of curved planar reformation images validates the clinical significance of the proposed framework. We release the code in \href{https://github.com/barry664/DeformCL}{https://github.com/barry664/DeformCL.}
\end{abstract}    
\section{Introduction}
\label{sec:intro}

Cerebro-cardiovascular disease is one of the most severe global health threats~\cite{roger2011heart}. Timely detection of potential high-risk locations and prompt intervention are crucial to reducing mortality, relying on accurate extraction, reconstruction, and analysis of major vessels in CTA images~\cite{leipsic2014scct}. In clinical practice, the primary requirement is to achieve accurate segmentation of primary vessel branches to enable comprehensive vascular system assessment and detailed local analysis. 

\begin{figure*}[t]
\centering
\includegraphics[width=1.0\linewidth]{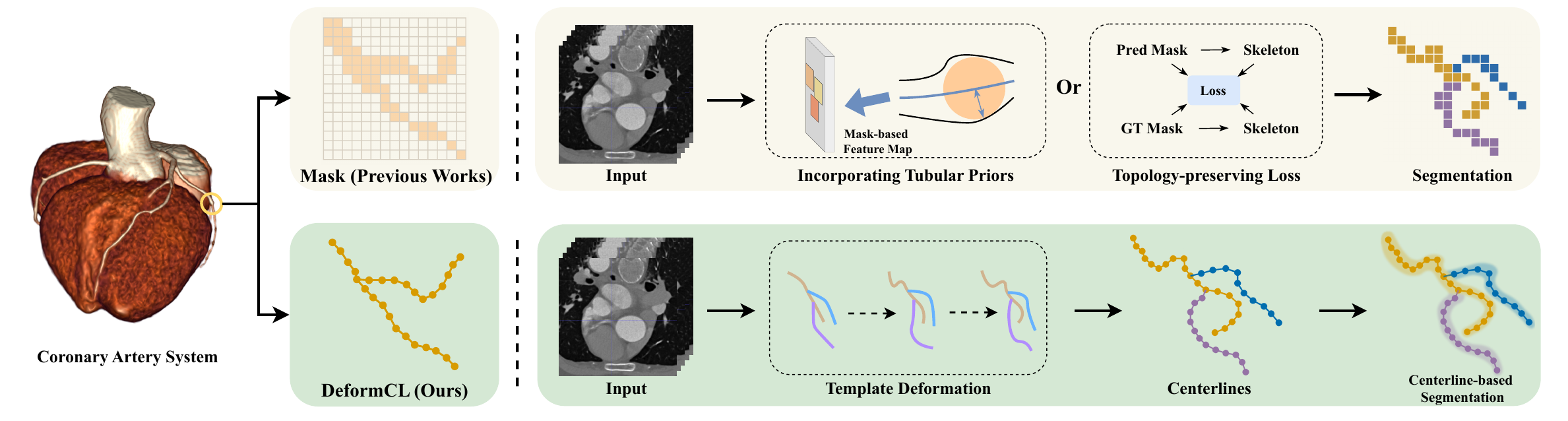}
\caption{Previous methods~\cite{qi2023dynamic, wang2020deep, shit2021cldice, araujo2021topological,wang2022pointscatter} primarily rely on the discrete mask representation, adopting a per-pixel classification paradigm for vessel segmentation. Common techniques in these approaches include incorporating tubular priors and applying topology-preserving loss functions. In contrast, we introduce a new continuous representation, \textbf{DeformCL}, which employs centerline points connected by edges to gradually fit the curvilinear vessels via a template deformation process. Once structured centerlines are obtained, generating segmentation predictions with improved topological accuracy becomes a straightforward task.}
\label{fig:intro}
%\vspace{-3pt}
\end{figure*}

Previous methodologies~\cite{qi2023dynamic, wang2020deep, shin2019deep,shit2021cldice, araujo2021topological,wang2022pointscatter} have predominantly adopted the traditional semantic segmentation approach, which relies on fine-grained, per-pixel classification within a discrete mask representation. As illustrated in the top part of Figure~\ref{fig:intro}, to improve the volumetric scores (\eg Dice) or topological metrics (\eg betti number errors) for segmentation, these approaches introduce a variety of innovations, such as incorporating tubular priors into network architecture~\cite{qi2023dynamic, wang2020deep, shin2019deep, zhang2022progressive, zhao2022graph, kong2020learning} or designing topology-preserving loss functions~\cite{shit2021cldice, araujo2021topological, hu2019topology}. Despite these advancements in the training process, the core issue that the \textbf{discrete mask representation is not well-suited for the curvilinear vessels} remains unchanged, imposing several inherent limitations that restrict the model's performance and applicability, as shown in the top part of Figure~\ref{fig:property}. Firstly, the neural networks struggle to recognize fragile and tortuous local structures, which may lead to segmentation fractures~\cite{qi2023dynamic}. Moreover, the discrete representation exhibits poor robustness to local noise~\cite{shit2021cldice, wang2022pointscatter}, such as variations in tissue density, due to the lack of a holistic view of the global morphology. Consequently, it may mistakenly identify the tissues with appearances similar to target vessels. Finally, designing an effective feature aggregation strategy along tubular curves remains challenging, considering the lack of explicit relationships among elements in discrete mask representations.

These challenges motivate us to seek a continuous representation that can explicitly fit the tubular structures from a holistic view, rather than confined to per-pixel classification paradigm. In this paper, we introduce DeformCL, a new representation for vessel extraction in 3D medical image based on \textbf{Deform}able \textbf{C}enter\textbf{l}ine. As shown in the bottom part of Figure~\ref{fig:intro}, DeformCL employs a graph structure, wherein centerline points serve as nodes connected by edges representing their spatial relationships. This continuously defined representation accurately depicts the curvilinear structures, thus circumventing the drawbacks of discrete mask representation and has its own merits: (1) the incorporation of natural connectivity mitigates centerline ruptures, particularly in the thin and fragile areas;
(2) by considering the entire curvilinear structures rather than local pixels, DeformCL ensures a comprehensive understanding of the global morphology, thus reducing local artifacts and false positives;
(3) the inherent graph structure facilitates efficient and effective feature aggregation along the tubular curves, enhancing overall segmentation performance.

We propose a novel framework to learn deformable centerline representation from 3D medical images. Our framework consists of several key steps. Firstly, we adaptively generate an initial continuous centerline template based on the input image, taking anatomical variations among different individuals into account. Next, the centerline points within the template will interact with each other to capture features along tubular curves. The whole centerline template will be iteratively refined by deformation and unpooling layers, gradually aligning with the ground truth centerlines.  Finally, after obtaining precise centerlines of tubular structures, crafting a high-quality segmentation algorithm becomes a straightforward task. In our approach, we leverage the distance map derived from the predicted centerline points to execute the final segmentation process effortlessly. Benefiting from the high-quality continuous centerlines, the segmentation performance is improved significantly.

We benchmark DeformCL on four 3D medical datasets, and the extensive experiments validate the effectiveness of our proposed method. Additionally, due to its continuous nature, DeformCL’s centerline predictions can be directly used as a structural representation of vessels, commonly as directional centerlines~\cite{schaap2009standardized, izzo2018vascular}. This structural representation is crucial in clinical practice, serving as an indispensable component in various subsequent diagnostic tasks, such as curved planar reformation~\cite{kanitsar2002cpr} and computational hemodynamics~\cite{taylor2010image}. We qualitatively demonstrate the clinical relevance of DeformCL, highlighting its potential applicability for diagnostic purposes.

In summary, our main contributions are three-fold:
\begin{itemize}
    \item We introduce a continuous representation, called DeformCL, which possesses several favorable properties for curvilinear vessels in 3D medical image.
    \item We propose a comprehensive pipeline for effectively training DeformCL through a cascaded deformation process.
    \item Our approach significantly bolsters the performance of several baselines on four 3D vessel extraction datasets.
\end{itemize}

\section{Related Work}
\subsection{Vessel Segmentation in 3D Medical Image}
Vessel segmentation in 3D medical image is a crucial task due to the prevalence of curvilinear vessels in the human body. Classical modern segmentation models like UNet~\cite{ronneberger2015u, cciccek20163d} and FCN~\cite{long2015fully} face challenges in accurately extracting tubular structures due to their complex global morphology and fragile thin structures. Previous methods have adopted various strategies to address these challenges: incorporate tubular priors into network architecture~\cite{qi2023dynamic, wang2020deep, shin2019deep, zhang2022progressive, zhao2022graph, kong2021deep, wu2024deep}, design topology-preserving loss functions~\cite{shit2021cldice, araujo2021topological, hu2019topology} or employ more flexible representation~\cite{wang2022pointscatter, zhang2024graphmorph}. Shit~\etal~\cite{shit2021cldice} introduces a new connectivity-aware similarity measure called clDice, and the differentiable version can be used to train segmentation models for tubular structures. Qi~\etal~\cite{qi2023dynamic} proposes a dynamic snake convolution to let networks adaptively focus on the tubular structures and make more accurate predictions. Wang~\etal~\cite{wang2022pointscatter} relies on the flexibility of point set representation which is not restricted to a fixed grid to promote the performance of tubular structure extraction.

\subsection{Deformable Paradigm}
The deformable paradigm is a widely used technique in deep learning, particularly in computer vision, where it plays a crucial role in capturing complex structures and achieving significant advancements in various tasks. Introduced by Dai et al.~\cite{dai2017deformable}, deformable convolution extends traditional convolutions to capture complex structures effectively. It has been successfully applied in tasks such as object detection and segmentation, allowing models to adapt to object variations and complex shapes. Deformable DETR~\cite{zhu2020deformable} combines attention mechanism with deformable paradigm for efficient feature extraction process. In addition, deformable structures have also found applications in 3D reconstruction tasks~\cite{wang2018pixel2mesh, wickramasinghe2020voxel2mesh, bongratz2022vox2cortex, kong2021deep, yang2022implicitatlas, zhao20213d}. For instance, Pixel2Mesh~\cite{wang2018pixel2mesh} generates 3D shapes represented as triangular meshes from single images by progressively deforming an initial ellipsoid. Similarly, Voxel2Mesh~\cite{wickramasinghe2020voxel2mesh} extends this approach to generate organ surfaces from 3D medical volumes, facilitating organ segmentation and visualization. Kong~\etal~\cite{kong2021deep} proposed a deep-learning approach to generate whole heart geometries from pre-defined mesh templates, demonstrating the versatility of deformable structures in human anatomical modeling. 

\section{Method}

In this section, we will first introduce the new representation DeformCL and its favorable properties in Sec~\ref{Sec: 3.1}. Sec~\ref{Sec: 3.2} elaborates on the network architecture for learning deformation centerline representation from 3D medical images. Sec~\ref{Sec: 3.3} presents the training losses for the framework.

\subsection{Representation of DeformCL}
\label{Sec: 3.1}

\begin{figure}[t]
\centering
\includegraphics[width=0.9\linewidth]{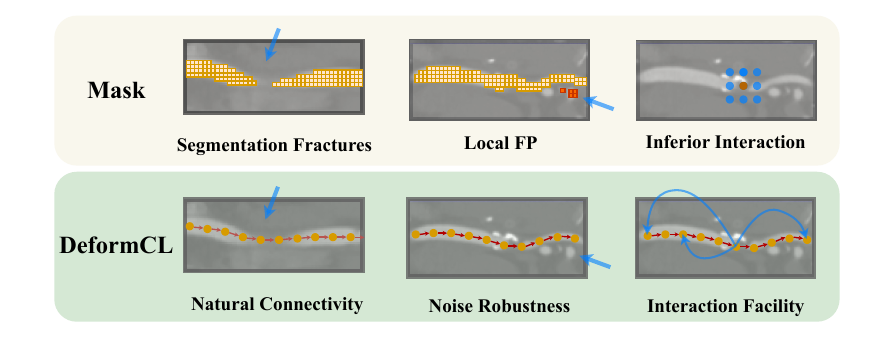}
\caption{Three significant properties of DeformCL compared with the discrete mask representation. (Zoom in for more details.)}
\label{fig:property}
\end{figure}

In 3D medical imaging, a common representation of a given volume $I \in \mathbb{R^{H \times W \times D}}$ is its segmentation mask, denoted as $Y \in \mathbb{R^{H \times W \times D}}$. Nearly all previous methods~\cite{qi2023dynamic, wang2020deep, shin2019deep,shit2021cldice, araujo2021topological,wang2022pointscatter}  rely on this discrete mask representation, framing the task as per-pixel classification. This paradigm inherently possesses several limitations, as illustrated in the top part of Figure~\ref{fig:property}, which ultimately restrict the model's performance.

In this paper, we introduce a novel continuous representation called \textbf{Deform}able \textbf{C}enter\textbf{L}ine (DeformCL) to efficiently and effectively capture tubular structures in 3D medical image. Instead of relying on segmentation mask, we represent tubular structures using a graph $\mathcal{G}=\{\mathcal{V}, \mathcal{\epsilon}\}$ , where $\mathcal{V}=\{v_i | v_i \in \mathbb{R}^3\}_{i=1}^N$ is a set of $N$ vertices standing for centerline points, $\mathcal{\epsilon} =\{(k, w)|1\leq k, w \leq N\} $ is the set of edges connecting adjacent vertices.

As shown in the bottom part of Figure~\ref{fig:property}, compared with previous discrete representations, DeformCL possesses several favorable properties, making it a reasonable and effective representation for tubular structure modeling.

\noindent\textbf{Natural connectivity.} Unlike traditional discrete representations, DeformCL naturally exhibits the connectivity among centerline points, which explicitly models the tubular structures in a continuously defined way. On one hand, this continuous representation helps prevent predictions from breaking in thin and fragile regions. On the other hand, the traditionally complex and time-consuming continuous centerlines extraction~\cite{izzo2018vascular, piccinelli2009framework} from segmentation for curved planner reconstruction~\cite{kanitsar2002cpr} is eliminated.

\begin{figure*}[t]
\centering
\includegraphics[width=0.9\linewidth]{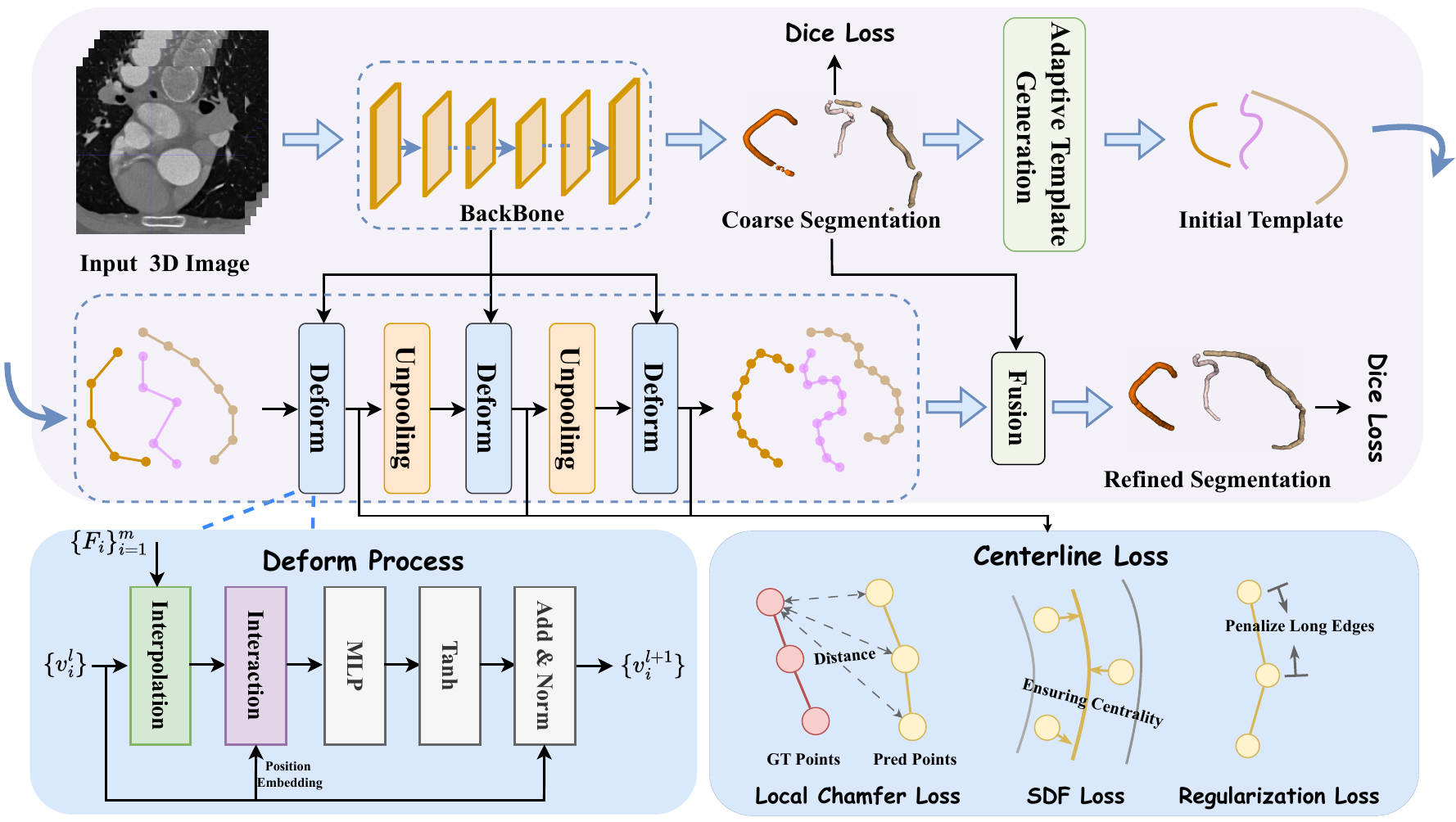}
\caption{Overview of the training process. 
Given a 3D medical image, the segmentation network first outputs a coarse prediction and generates hierarchical downsampled feature maps. An initial template, produced by the Adaptive Template Generation module, is then iteratively fed through deformation and unpooling layers to progressively adapt to the curvilinear structures. Finally, refined segmentation results are generated based on the accurately aligned continuous centerlines. We apply Dice loss to both the coarse and refined segmentation predictions, and local chamfer loss, SDF loss, and regularization loss for centerline predictions.}
\label{fig: model}
\end{figure*}

\noindent\textbf{Noise robustness.} Thanks to its top-down manner, DeformCL exhibits robustness to noise and local disturbances, such as variations in tissue density or the presence of contrast agents. This inherent property enables DeformCL to effectively filter out local artifacts and false positives, leading to more accurate and reliable tubular structure extraction compared to previous bottom-up methods.

\noindent\textbf{Interaction facility.} Constrained by the discrete representation, previous methods~\cite{shit2021cldice,wang2022pointscatter,qi2023dynamic,wang2020deep} lack of an effective approach for tubular feature aggregation. On the contrary, DeformCL's graph structure facilitates feature fusion process via methods like GCN~\cite{kipf2016semi} or Transformer~\cite{vaswani2017attention} along the tubular curve. This global information interaction is  significant for the structure completeness and topology correctness of tubular predictions.

\subsection{Network Architecture}
\label{Sec: 3.2}

In this subsection, we will introduce the network architecture for learning deformable centerline representation. The whole framework is shown in Figure~\ref{fig: model}. Given a 3D medical image $I \in \mathbb{R^{H\times W \times D}}$, we first pass it through a 3D backbone (\eg 3D UNet~\cite{cciccek20163d}) to obtain hierarchical feature maps $\{F_i\}_{i=1}^m$ at different down-sampled rates. Additionally, the backbone network produces a coarse segmentation result $\hat{Y}_c \in \mathbb{R^{H\times W \times D}}$ for each category of vessels.

\noindent\textbf{Adaptive Template Generation.} %In order to facilitate the centerline deformation process, we employ an adaptive template generation approach which encodes the input image information to a certain extent.
In order to initialize the graph structure of DeformCL, we attempt to initialize a tubular template motivated by mesh generation methods~\cite{wang2018pixel2mesh, wickramasinghe2020voxel2mesh}.
However, unlike human organs, tubular structures in 3D medical image, such as coronary arteries, exhibit complex global morphology and variable extension direction, which means that discovering a universal template for such structures poses significant challenges.
Therefore, we introduce a new method to adaptively generate initial templates for each image.

\begin{figure}
\centering
\includegraphics[width=1.0\linewidth]{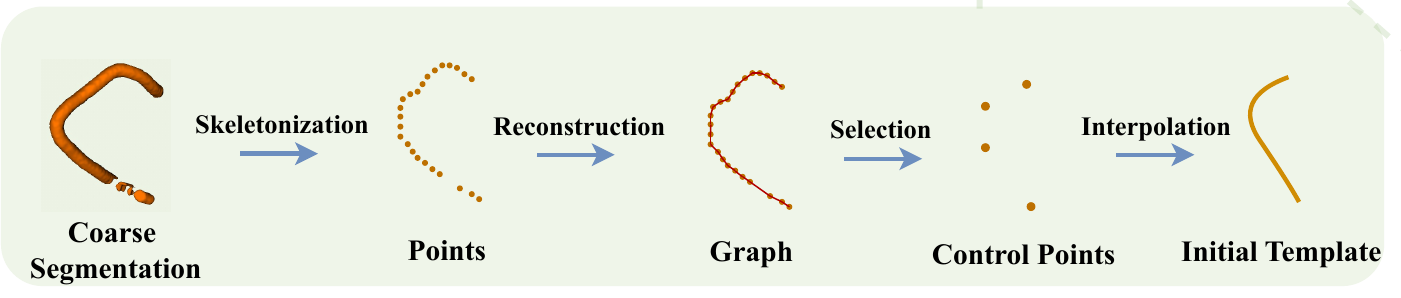}
\caption{Adaptive Template Generation.}
\label{fig: temp}
\end{figure}

The template generation process, illustrated in the Figure~\ref{fig: temp}, is conducted independently for each category of vessels. %, and we assume that each category of vessels does not have branches. 
Firstly, the coarse segmentation result $\hat{Y}_c$ is converted into discrete centerline points via traditional 3D thinning algorithm~\cite{lee1994building}. These points are then reconstructed into a graph using the minimum spanning tree algorithm~\cite{graham1985history} following the strategy in~\cite{zhang2023topology}. Next, a set of $k$ control points is selected from the graph at fixed distances, serving as the abstraction of the structure. Finally, curvilinear interpolation (\eg Bezier Curve Interpolation, Linear Interpolation) is applied based on the control points to generate the initial template, denoted as $\{v_i^c |v_i^c \in \mathbb{R}^3\}_{i=1}^{N_c}$. Here,  $v_i^c$ represents the $i$-th point for category $c$, $N_c$ is the total number of points for category $c$, and $v_i^c$ is connected with $v_{i+1}^c$ by default.

It is worth noting that we adopt a rough template generation strategy using control points instead of directly utilizing the reconstructed tree as the initial template. The reason is that we empirically find overly accurate initial templates may harm model performance, supported by experimental results shown in Table~\ref{tab: ablation temp}. We speculate that overly accurate initial templates may simplify the deformation task excessively, potentially leading to insufficient model training.

\noindent\textbf{Cascaded Deformation Process.} After the generation of initial templates, we proceed with the cascaded deformation process to refine the centerline predictions. For clarity, we focus on the centerline points of one category of vessels in the subsequent sections of the paper.

Firstly, trilinear interpolation is applied to each centerline point $v_i$ in the downsampled feature maps $\{F_i\}_{i=1}^m$, and the hierarchical points features are fused via a linear layer:
\begin{equation}
    f(v_i) = \text{Linear}([\text{Tri}(F_1, v_i), ..., \text{Tri}(F_m, v_i)]) \in \mathbb{R}^C,
\end{equation}
where $C$ denotes the channel size, Tri represents the trilinear interpolation. 
%For convenience we take the centerline points of one category of vessels for example in the following parts of the paper.
Next, the points and their corresponding features $\{v_i, f(v_i)\}$ are fed into Transformer~\cite{vaswani2017attention} layers
 for feature aggregation along the tubular curve. The positional encoding $\text{Pos}(v^l)$ contains learnable embeddings of absolute 3D positions of points and relative positions in the graph. Subsequently, the augmented point features are used to predict an offset $\delta_i^{l}$, which is then applied to update the position of each point $v_i^{l}$ to obtain $v_i^{l+1}$. 
 
Since high-accuracy centerlines require a sufficient number of points to capture curvilinear structures accurately, we apply an unpooling strategy after deformation to increase point density. Specifically, a new point is inserted between each pair of connected points. We reuse the symbol $\{v_i^{l+1}\}$ for clarity. This process is iterated $L$ times for accurate centerline prediction, where the $l$-th layer is described as follows:
 \begin{align}
     &\tilde{f}(v_i^l) = f(v_i^l) + \text{Transformer}(f(v^l), \text{Pos}(v^l)) \in \mathbb{R}^C, \\
     &\delta_i^{l} = \text{Tanh}(\text{MLP}(\tilde{f}(v_i^{l}))) \in \mathbb{R}^3, \\
     &v_i^{l+1} = v_i^l + \delta_i^{l} \in \mathbb{R}^3, \\
     &\{v_i^{l+1}\} = \text{Unpooling}(\{v_i^{l+1}\})
 \end{align}
After completing the cascaded deformation process, the final set of points $\{v_i^L\}$ represents the predicted centerline points, which will be used for subsequent centerline-based segmentation task.

\noindent\textbf{Centerline-based Segmentation.} Once accurate centerlines of tubular structures are obtained, predicting the corresponding segmentation results becomes straightforward. In this work, we employ a simple method using distance maps to achieve centerline-based segmentation. Specifically, given the predicted centerline points $V=\{v_i^L \in \mathbb{R}^3\}$, we calculate the minimum distance of each voxel in the image to the points in $V$:

\begin{equation}
    D[x, y, z] = \mathop{\min}\limits_{v \in V} || (x, y, z) - v ||_2,
\end{equation}

where $D \in \mathbb{R^{H \times D \times W}}$ is the distance map based on the centerline points. Finally, the coarse segmentation results $\hat{Y}_c$ and the distance map $D$ are input to a convolutional layer for fusion, yielding the refined segmentation results $\hat{Y}_r$.

\subsection{Training Loss}
\label{Sec: 3.3}

In this subsection, we will elaborate on the loss functions of the proposed framework. For centerline prediction, we employ a combination of local chamfer distance loss, signed distance function (SDF) loss, and a regularization loss. We utilize Dice loss for segmentation.

\noindent\textbf{Local Chamfer Distance Loss.}
Tradition global chamfer distance loss calculates the distance between two point cloud sets. However, curvilinear vessels in the human body often exhibit complex global morphology and sparse 3D distribution. As a result, using global chamfer distance loss may lead to larger fluctuations and suboptimal convergence during the training process. To address this issue, we adopt a local version of chamfer distance loss. 

To be specific, we begin by employing Farthest Point Sampling (FPS) to select candidate points $\{u_{i_k}\}$ from the ground truth centerline points $\{u_i\}$, which are obtained from the ground truth mask via traditional 3D thinning algorithm~\cite{lee1994building}. Then, we crop local patches with fixed size $M$ centered around each selected point  $\{u_{i_k}\}$. The points located within each cropped patch are denoted as $\omega_{i_k} = \{p=(x, y, z) | p \in [u_{i_k} - M / 2, u_{i_k} + M / 2 ]\}$, and the set of patches is denoted as $\Omega = \{\omega_{i_k}\}$. The chamfer distance loss is then applied between the predicted point set and the ground truth point set within each cropped patch. Formally, the loss is defined as:

\begin{equation}
    \mathcal{L}_{\text{cha}} = \frac{1}{|\Omega|}\sum_{\omega \in \Omega} (\sum_{v_i \in \omega} \mathop{\min}\limits_{u_i \in \omega} || v_i - u_i ||_2^2 + \sum_{u_i \in \omega} \mathop{\min}\limits_{v_i \in \omega} || u_i - v_i ||_2^2).
\end{equation}

\noindent\textbf{SDF Loss.}
Signed Distance Function (SDF) is a well-established concept in computer vision and computer graphics, used to represent the surface of 3D objects through signed distances. We find that SDF is appropriate for ensuring the centrality of predicted centerline points.
Firstly, we define the surface of tubular strcutures following~\cite{wang2020deep}:
\begin{equation}
    S = \{x \in \mathbb{R}^3 | Y[x]=1, \exists u \in \mathcal{N}(x), Y[u]=0 \},
\end{equation}
where $\mathcal{N}(x)$ denotes the 6-neightbour voxels of $x$, $Y$ is the ground truth mask. Then, the SDF value of an arbitrary point p in the image can be calculated as:
\begin{align}
SDF(p) =
\left\{
             \begin{array}{cl}
             -\mathop{\min}\limits_{u \in S} || p - u ||_2^2, & \text{if} \ Y[p]=1,  \\
             \mathop{\min}\limits_{u \in S} || p - u ||_2^2, &   \text{if} \ Y[p]=0.    \\
             \end{array}
\right. 
\end{align}

\begin{comment}
\begin{figure}
\centering
\includegraphics[width=0.8\linewidth]{images/sdf_loss.pdf}
\caption{SDF Loss.}
\label{fig: SDF Loss}
\end{figure}
\end{comment}

The SDF values of all predicted centerline points are utilized as the training loss:
\begin{equation}
    \mathcal{L}_{\text{sdf}} = \frac{1}{N}\sum_{i=1}^{N} SDF(v_i).
\end{equation}

\begin{wrapfigure}[8]{r}{0.25\textwidth}
\centering
\vspace{-12pt}
\includegraphics[width=1.0\linewidth]{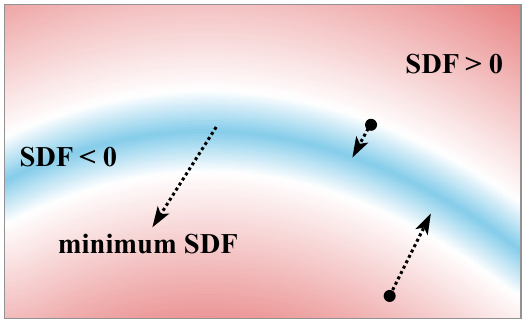}
%\vspace{-10pt}
\caption{SDF Loss.}
\label{fig: SDF Loss}
\end{wrapfigure}

As illustrated in Figure~\ref{fig: SDF Loss},
SDF loss serves two crucial purposes. Firstly, it provides supervision for all predicted centerline points, whereas the local chamfer distance loss can only provide supervision for points within cropped patches, potentially missing outlier predictions. Secondly, it facilitates the gradual deformation of predicted points towards the ideal center of tubular structures, thereby ensuring the centrality of centerlines.

\noindent\textbf{Regularization Loss.}
To prevent the deformable centerline points from deviating mistakenly, which may result in long edges between adjacent vertices, we apply a regularization loss to penalize the adjacent points with much long edges:
\begin{equation}
    \mathcal{L}_{\text{reg}} = \frac{1}{N} \sum_{i=1}^N ||v_{i} - v_{i-1}||_2^2.
\end{equation}
\noindent\textbf{Segmentation Loss.} We use Dice loss~\cite{milletari2016v} as loss function for the coarse and refined segmentation results $\hat{Y}_c$ and $\hat{Y}_r$:
\begin{equation}
    \mathcal{L}_{\text{seg}} = \lambda \text{Dice}(Y, \hat{Y}_c) + \text{Dice}(Y, \hat{Y}_r),
\end{equation}
where $\lambda$ is set to 0.2.

Note that local chamfer distance loss, SDF loss and regularization loss are applied to all intermediate results in the cascaded deformation process.
The total training loss is a combination of all loss functions:
\begin{equation}
    \mathcal{L} = \sum_{l=1}^L{w_{l}(\lambda_{\text{cha}}\mathcal{L}_{\text{cha}} + \lambda_{\text{sdf}}\mathcal{L}_{\text{sdf}} + \lambda_{\text{reg}}\mathcal{L}_{\text{reg}})
    } + \mathcal{L}_{\text{seg}},
\end{equation}
$w_l$ stands for the weight of the $l$-th deformation stage.
\section{Experiments}

\subsection{Datasets}
We use four datasets to evaluate our proposed framework. \textbf{HaN-Seg}~\cite{podobnik2023han} is a public dataset consisting of 56 CT and MR images with 30 categories of head and neck organ pixel annotations. 
\textbf{ASOCA}~\cite{gharleghi2022automated, gharleghi2023annotated} is a public dataset in the MICCAI-2020 challenge which aims to automatically segment the coronary artery lumen. It consists of 40 Cardiac Computed Tomography Angiography (CCTA) images with artery segmentation annotations. \textbf{ImageCAS}~\cite{zeng2023imagecas} is a public dataset containing about 1000 3D CTA images with coronary artery segmentation labels. \textbf{HNCTA} is a private head and neck artery dataset with 358 CTA images. More details of datasets can be found in supplementary materials.

\subsection{Evaluation Metrics}
We assess the segmentation performance in three main aspects following~\cite{qi2023dynamic}. For volumetric scores, we report Dice~\cite{milletari2016v} and clDice~\cite{shit2021cldice} to evaluate the segmentation results. For topology-based scores, absolute Betti Errors for Betti Numbers $\beta_0$ and $\beta_1$ and absolute error of Euler characteristic are used to measure the topology and completeness of the extracted vessels following~\cite{shit2021cldice, wang2022pointscatter, qi2023dynamic}. For distance-based scores, Hausdorff Distance (HD) is employed for measuring the maximum distance between two 3D point sets. All presented metrics in this paper are the average of all vessel categories.

For the centerline extraction task, we use F1, Hausdorff Distance, Chamfer Distance and absolute Betti Errors for Betti Numbers $\beta_0$ as evaluation metrics. For robust evaluation, we apply a three-pixel tolerance region around the ground truth centerline points following~\cite{wang2022pointscatter, guimaraes2016fast}.

\begin{table}[t]
\centering
\caption{Main results on vessel segmentation task. We mark the best performance by bold numbers. \textbf{Ours} (softDice) denotes that we use softDice as base model with the proposed DeformCL as an additional segmentation head.} 
\label{tab:seg}
\renewcommand\arraystretch{1.2}
\scalebox{0.59}{
\begin{tabular}{c | l | c c | c c c | c}
\toprule
\mrtwo{Dataset}	&\mrtwo{Method}	& \multicolumn{2}{c|}{Volumetric Scores (\%) $\uparrow$}& \multicolumn{3}{c|}{Topological Error $\downarrow$} & Distance $\downarrow$ \\
\cline{3-8}
&	&Dice	&clDice	&$\beta_0$	&$\beta_1$	& Euler & HD95	\\
\midrule										
    & DDT~\cite{wang2020deep} & 75.76 & 85.79  & 2.333 & 0.127 & 2.329 & 5.473 \\
 	& softDice~\cite{milletari2016v} & 76.30 & 85.96  & 3.316 & \textbf{0.029} & 3.286 & 4.905 \\
    \rowcolor{mygray}\cellcolor{white}{}&\textbf{Ours} (softDice) & $\textbf{79.58}_{+3.28}$ & $\textbf{88.86}_{+2.90}$  & \textbf{2.123} & 0.089 & \textbf{2.034} & 4.113 \\
    {HaN-Seg}&	clDice~\cite{shit2021cldice}	& 76.22 & 86.50  & 2.861 & 0.065 & 2.796 & 5.244 \\
 \rowcolor{mygray}\cellcolor{white}{}&\textbf{Ours} (clDice) & $78.39_{+2.17}$ & $87.32_{+0.82}$ & 2.207 & 0.226        & 2.042      & \textbf{3.838} \\
    & DSCNet~\cite{qi2023dynamic} & 77.16 & 84.63  & 3.858 & 0.295 & 3.564 & 4.949 \\
    \rowcolor{mygray}\cellcolor{white}{}&\textbf{Ours} (DSCNet) & $78.61_{+1.45}$          & $86.89_{+2.26}$          & 2.602          & 0.358 & 2.318          & 4.125 \\

\midrule										
    & DDT~\cite{wang2020deep} & 83.56          & 93.28          & 0.172          & 0.050          & 0.219 & 2.666          \\
	&softDice~\cite{milletari2016v} & 83.05          & 92.14          & 0.286          & 0.059          & 0.337          & 2.924          \\
    \rowcolor{mygray}\cellcolor{white}{}&\textbf{Ours} (softDice) & $84.69_{+1.64}$ & $93.72_{+1.58}$ & \textbf{0.149} & 0.051        & \textbf{0.199}       & \textbf{2.084} \\
    {HNCTA}& clDice~\cite{shit2021cldice}	& 83.13          & 91.89         & 0.160        & 0.058          & 1.239          & 3.118         \\
 \rowcolor{mygray}\cellcolor{white}{}&\textbf{Ours} (clDice) & $84.13_{+1.00}$ & $93.15_{+1.26}$ & 0.150 & 0.052       & 0.200      & 2.839 \\
    & DSCNet~\cite{qi2023dynamic} & 83.54          & 92.06          & 0.246          & 0.060          & 0.301          & 2.658          \\
    \rowcolor{mygray}\cellcolor{white}{}&\textbf{Ours} (DSCNet) & $\textbf{84.80}_{+1.26}$          & $\textbf{93.82}_{+1.76}$          & 0.183 & \textbf{0.044} & 0.228          & 2.349 \\
\midrule										
    & DDT~\cite{wang2020deep} & 83.00 & 89.63  & 0.998 & 0.094 & 1.090 & 7.646 \\
    &softDice~\cite{milletari2016v} & 84.06 & 89.81  & 0.930 & 0.119 & 0.996 & 6.829 \\
    \rowcolor{mygray}\cellcolor{white}{}&\textbf{Ours} (softDice) & $85.16_{+1.10}$ & $92.41_{+2.60}$  & \textbf{0.408} & 0.129 & \textbf{0.510} & 5.082 \\
    {ASOCA}& clDice~\cite{shit2021cldice}	& 84.08 & 90.92  & 1.009 & 0.122 & 1.082 & 6.299 \\
 \rowcolor{mygray}\cellcolor{white}{}&\textbf{Ours} (clDice) & $84.77_{+0.69}$ & $91.94_{+1.02}$ & 0.603 & 0.100       & 0.677       & \textbf{4.122} \\
    & DSCNet~\cite{qi2023dynamic} & 84.28 & 91.04  & 0.840 & 0.146 & 0.927 & 5.286\\
    \rowcolor{mygray}\cellcolor{white}{}&\textbf{Ours} (DSCNet)& $\textbf{85.47}_{+1.19}$ & $\textbf{92.87}_{+1.83}$  & 0.508 & \textbf{0.083} & 0.590 & 5.212 \\
\midrule										
    &   DDT~\cite{wang2020deep} & 87.51 & 95.32  & 0.189 & 0.065 & 0.278 & 4.621 \\ 
 	&softDice~\cite{milletari2016v} & 87.59 & 95.15  & 0.190 & \textbf{0.062} & 0.268 & 4.674 \\ 
    \rowcolor{mygray}\cellcolor{white}{}&\textbf{Ours} (softDice) & $88.33_{+0.74}$ & $95.48_{+0.33}$  & \textbf{0.074} & 0.065 & \textbf{0.159} & 3.580 \\ 
    {ImageCAS}&	clDice~\cite{shit2021cldice}	& 87.78 & 96.00  & 0.158 & 0.064 & 0.238 & 4.021 \\
  \rowcolor{mygray}\cellcolor{white}{}&\textbf{Ours} (clDice) & $87.85_{+0.07}$ & $96.01_{+0.01}$ & 0.152 & 0.067       & 0.234      & 5.038 \\
    & DSCNet~\cite{qi2023dynamic} & 87.98 & 95.34  & 0.213 & 0.070 & 0.283 & 3.751 \\
    \rowcolor{mygray}\cellcolor{white}{}&\textbf{Ours} (DSCNet) & $\textbf{88.84}_{+0.86}$ & $\textbf{96.38}_{+1.04}$  & 0.127 & 0.069 & 0.215 & \textbf{3.291}	\\

\bottomrule										
\end{tabular}
}
\end{table}

\begin{table}
\centering
\caption{Main results on centerline extraction task.} 
\label{tab:cl}
\renewcommand\arraystretch{1.2}
\scalebox{0.59}{
\begin{tabular}{c | l | c c c c}
\toprule

\mrtwo{Dataset}	&\mrtwo{Method}	& \multicolumn{4}{c}{Metrics} \\
\cline{3-6}
&	&F1 (\%)$\uparrow$	 & HD95 $\downarrow$	& Chamfer $\downarrow$	&  $\beta_0$ $\downarrow$	\\
\midrule										
    & DDT~\cite{wang2020deep} & 89.65   & 7.678 & 18.957 & 3.136  \\
    
 	& softDice~\cite{milletari2016v} & 91.17  & 6.981 & 19.111 & 4.318 \\
  
  \rowcolor{mygray}\cellcolor{white}{}&\textbf{Ours} (softDice) & $\textbf{94.56}_{+3.39}$  & \textbf{4.250} & \textbf{6.332} & 1.727 \\
  
{HaN-Seg} & clDice~\cite{shit2021cldice}	& 90.78 & 9.023 & 29.217 & 3.636  \\
    
 \rowcolor{mygray}\cellcolor{white}{}&\textbf{Ours} (clDice) & $92.36_{+1.58}$ & 5.220 & 7.130    & \textbf{1.409}  \\
 
    & DSCNet~\cite{qi2023dynamic} & 90.45 & 7.462 & 18.295 & 4.818 \\
    
    \rowcolor{mygray}\cellcolor{white}{}&\textbf{Ours} (DSCNet)& $91.50_{+1.05}$  & 8.222 & 12.362 &2.773\\

\midrule										
    & DDT~\cite{wang2020deep} & 97.29   & 3.291 & 7.083 & 0.394  \\
    
 	& softDice~\cite{milletari2016v} & 97.05  & 3.821 & 11.994 & 0.591 \\
  
  \rowcolor{mygray}\cellcolor{white}{}&\textbf{Ours} (softDice) & $\textbf{98.74}_{+1.69}$  & 2.250 & 4.072 & 0.422 \\
  
{HNCTA} & clDice~\cite{shit2021cldice}	& 96.69 & 3.566 & 9.097 & \textbf{0.387}  \\
    
 \rowcolor{mygray}\cellcolor{white}{}&\textbf{Ours} (clDice) & $98.36_{+1.67}$ & 2.321 & 6.761    & 0.520  \\
 
    & DSCNet~\cite{qi2023dynamic} & 97.28 & 3.397 & 8.405 & 0.534 \\
    
     \rowcolor{mygray}\cellcolor{white}{}&\textbf{Ours} (DSCNet)&  $98.59_{+1.31}$ & \textbf{2.145} & \textbf{3.829} & 0.419  \\

\midrule										
    & DDT~\cite{wang2020deep} & 91.43  & 10.641 & 63.487 & 2.063  \\
    
 	& softDice~\cite{milletari2016v} & 91.57 & 11.956 & 46.358 & 1.905 \\
  
  \rowcolor{mygray}\cellcolor{white}{}&\textbf{Ours} (softDice) & $\textbf{95.23}_{+3.66}$ & 7.423 & 29.027 & \textbf{0.810} \\
  
{ASOCA} & clDice~\cite{shit2021cldice}	& 93.43  & 9.168 & 29.971 & 1.554  \\
    
 \rowcolor{mygray}\cellcolor{white}{}&\textbf{Ours} (clDice) & $94.77_{+1.34}$ & 7.273 & 17.738  & 0.929  \\
 
    &DSCNet~\cite{qi2023dynamic} & 93.46 & 7.908 & 23.052 & 1.540 \\
    
    \rowcolor{mygray}\cellcolor{white}{}&\textbf{Ours} (DSCNet)& $94.28_{+0.82}$ & \textbf{7.052} & \textbf{14.760} & 1.000\\
\midrule										
    & DDT~\cite{wang2020deep} & 95.10  & 4.881 & 19.376 & 0.538  \\
    
 	& softDice~\cite{milletari2016v} & 94.40 & 7.134 & 30.345 & 0.687 \\
  
  \rowcolor{mygray}\cellcolor{white}{}&\textbf{Ours} (softDice) & $96.67_{+2.27}$ & 5.127 & 17.313 & \textbf{0.123} \\
  
{ImageCAS} & clDice~\cite{shit2021cldice}	& 95.78  & 5.043 & 19.490 & 0.385  \\
    
 \rowcolor{mygray}\cellcolor{white}{}&\textbf{Ours} (clDice) & $96.64_{+0.86}$ & 4.743 & 14.941  & 0.333  \\
 
    &DSCNet~\cite{qi2023dynamic} & 95.32 & 5.170 & 17.239 & 0.503 \\
    
    \rowcolor{mygray}\cellcolor{white}{}&\textbf{Ours} (DSCNet)& $\textbf{97.28}_{+1.96}$ & \textbf{4.303} & \textbf{12.152} & 0.318\\

\bottomrule										
\end{tabular}
}
\vspace{-14pt}
\end{table}

\subsection{Implementation Details}

In our experiments, we use 3D UNet~\cite{cciccek20163d} as the backbone for all methods. We set the number of hierarchical feature maps $m$ to 5. %which include features with $1\times, 4\times, 8\times, 16\times$, and $32\times$ down-sampling, respectively. 
The number of control points $k$ in template generation is set to 4.
The number of deformation stages  $L$ is 4, and the channel size for point features $C$ is 24.
The centerline template of each vessel category is initialized with $N_c = 100$ points by default.
All of the coordinates of centerline points are normalized to 0-1 in the framework.

When computing local chamfer distance loss, the number of local patches is sampled from an even distribution $|\Omega| \sim U(60, 80)$. The loss weights $\lambda_{\text{cha}}$, $\lambda_{\text{sdf}}$, $\lambda_{\text{reg}}$ are set to 30, 0.5, and 60 by default. The balance weights for different deformation stages $\{w_l\}_{l=1}^L$ are set to 0.05, 0.60, 0.95, and 1.00.
More implementation details are depicted in the supplementary materials.

\begin{figure}[t]
\centering
\includegraphics[width=1.0\linewidth]{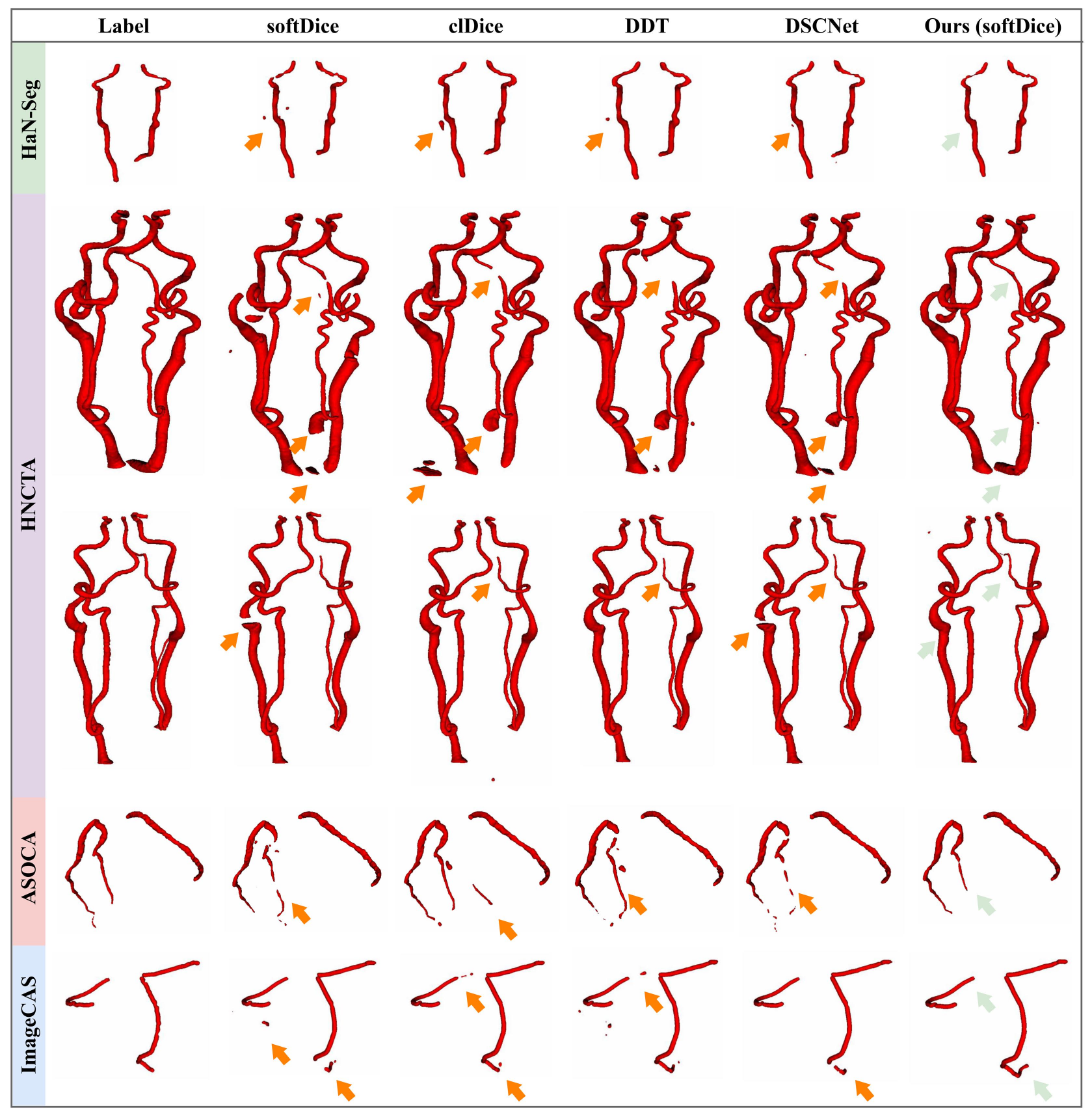}
\caption{Visualizations of the predictions from different methods on four 3D medical image datasets. Different rows and columns stand for different datasets and methods respectively. The arrows point out the areas where models easily exhibit errors. Zoom in for more details.}
\label{fig:visual}
\end{figure}

\subsection{Main Results}
Our approach can be incorporated into common segmentation models, such as softDice~\cite{milletari2016v} or clDice~\cite{shit2021cldice}, as an additional segmentation head.  In this subsection, we present the results of our method implemented on softDice~\cite{milletari2016v}, clDice~\cite{shit2021cldice} and DSCNet~\cite{qi2023dynamic}, comparing it with competitive segmentation models, including softDice~\cite{milletari2016v}, cldice~\cite{shit2021cldice}, DDT~\cite{wang2020deep} and DSCNet~\cite{qi2023dynamic}.
All models are trained under identical settings, including training schedules and hyper-parameters, ~\etc. We have not included PointScatter~\cite{wang2022pointscatter} and Topoloss~\cite{hu2019topology} in the comparison as it is designed for 2D images and is not applicable to 3D medical images due to GPU memory limitations.

For segmentation, our method demonstrates notable performance improvements when integrated with baseline models as depicted in Table~\ref{tab:seg}. The enhancement is particularly prominent in metrics like clDice and Betti number errors, which primarily reflect vessel completeness and topology correctness. This phenomenon verifies the effectiveness of the continuously defined representation. Furthermore, thanks to the exact prediction of centerlines, the performance of segmentation on the distance metric HD95 has also improved a lot.

For centerline extraction, we present the experimental results on HaN-Seg and ASOCA datasets in Table~\ref{tab:cl}. It can be concluded that DeformCL can improve a range of metrics consistently across multiple baseline models. More experiments can be found in supplementary materials.

To provide a qualitative assessment of the model performance, we present visualizations of predictions from different methods in Figure~\ref{fig:visual}. These results illustrate that DeformCL effectively captures thin and fragile structures with minimal segmentation fractures. Moreover, the reduction in scattered false positives demonstrates the noise robustness of DeformCL. These results reaffirm that the continuously defined representation offers superior capabilities in curvilinear vessel extraction, owing to its unique properties.

\begin{table}[t]
\begin{minipage}{0.48\textwidth}
\centering
\caption{Ablation on the number of control points k for each category of vessels on HaN-Seg dataset.} 
\label{tab: ablation temp}
\renewcommand\arraystretch{1.2}
\scalebox{0.79}{
\begin{tabular}{c|cc}
\toprule
 \multirow{2}{*}{k} & \multicolumn{2}{c}{Volumetric Scores (\%) $\uparrow$} \\ \cline{2-3} 
   & Dice         & clDice            \\ \hline
 0            & 78.87              & 86.89              \\
    \rowcolor{mygray} $4^*$       & \textbf{79.58}     & \textbf{88.86}   \\
 8                 & 78.36             & 86.88             \\
 16                   & 77.22              & 85.93              \\
64                  & 77.37              & 85.66              \\ \bottomrule
\end{tabular}
}
\end{minipage}
\hfil
\hfil
\begin{minipage}{0.48\textwidth}
\centering
\caption{Ablation on loss functions on HaN-Seg dataset. w/o $\mathcal{L}_{\text{cha}}$ means that we use global chamfer distance loss instead of local chamfer distance loss.} 
\label{tab: ablation loss}
\renewcommand\arraystretch{1.2}
\scalebox{0.79}{
\begin{tabular}{c|cc}
\toprule
\multirow{2}{*}{Loss Function}             & \multicolumn{2}{c}{Volumetric Scores (\%) $\uparrow$}               \\ \cline{2-3}                  
& \multicolumn{1}{c}{Dice} & \multicolumn{1}{c}{clDice} \\ \hline
 w/o $\mathcal{L}_{\text{cha}}$                              & 78.18                    & 86.62                      \\
 w/o $\mathcal{L}_{\text{reg}}$                                                  & 78.38                    & 86.70                      \\
 w/o $\mathcal{L}_{\text{sdf}}$                                                        & 79.51                   & 87.89                     \\
\rowcolor{mygray} full configuration            & \textbf{79.58}     & \textbf{88.86}   \\ \bottomrule
                         % & 85.87                                  & 78.18                    & 86.62                      \\
\end{tabular}
}
\end{minipage}
\end{table}

\subsection{Ablation Study}
\label{Sec: 4.5}
We conduct ablation experiments on the HaN-Seg dataset to assess the impact of different components of our method. Other ablation studies are presented in the supplementary materials.

\noindent\textbf{Template Selection.}
We compare the performance of different template selection strategies in Table~\ref{tab: ablation temp}. k=0 means that we use a straight centerline as the initial template without considering image information. It can be found that adaptive template generation outperforms fixed initial templates (k=0). Furthermore, we observe a gradual decline in performance as the number of control points $k$ increases from 4 to 64. We guess that more control points lead to overly accurate initial templates, thus reducing the difficulty and complexity of the deformation task. Consequently, the model may not be sufficiently trained. Therefore, we adopt a rough initial template with k=4.

\begin{figure}[t]
\centering
\includegraphics[width=0.9\linewidth]{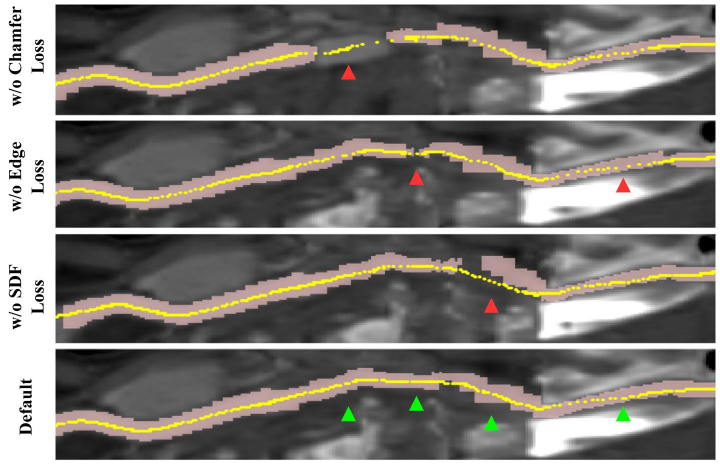}
\caption{Visual comparison of different loss function configurations. The \textcolor[rgb]{0.75,0.65,0.65}{carnatio} masks and \textcolor[rgb]{0.8 , 0.8, 0.35}{yellow} lines denote the predicted segmentations and centerlines respectively. The \textcolor{Red}{red} and \textcolor[rgb]{0.3,0.85,0}{green} triangles point out the areas to focus on.}
\label{fig: ablation}
\end{figure}

\noindent\textbf{Loss Function.}
We evaluate the effectiveness of different loss functions in Table~\ref{tab: ablation loss} and Figure~\ref{fig: ablation}. It is observed from the results in the table that local chamfer distance loss outperforms global chamfer distance loss, particularly in terms of clDice. We empirically find that the global chamfer distance loss will result in larger fluctuations during training process compared with the local version. The reason is that the tubular structures in human body usually have complex global morphology and sparse 3D distribution, which means that conducting global calculations will have more fluctuations. 

Although the gains from SDF loss and regularization loss are not as pronounced in the numerical results, visualization comparison in the figure reveal that these losses contribute to more accurate centerline predictions. For instance, without regularization loss, the centerline prediction may be sparse and deviated in local regions, due to the long edges between adjacent vertices. In addition, adding SDF loss augments the centrality of centerline predictions. These properties are significant for the subsequent tasks, such as curved planar reformation~\cite{kanitsar2002cpr}. In summary, the full configurations of loss functions achieve the best performance.

\subsection{Clinical Significance}
\label{Sec: 4.6}

In clinical practice, the generation of straightened curved planar reformation~\cite{kanitsar2002cpr} (SCPR) images is vital for radiologists to conduct accurate diagnoses for blood vessels, especially for observing possible plaques and stenosis. High-quality SCPR images require high-quality vessel graphical structures (directional centerlines) that meet the criteria of completeness, smoothness, and centrality. Previous methods often rely on converting discrete segmentation masks into continuous centerlines using traditional tools~\cite{izzo2018vascular,piccinelli2009framework}, which may be susceptible to imperfect segmentation predictions. Moreover, this conversion process is complex and time-consuming. In contrast, our DeformCL can directly generate continuous centerlines during model inference without the need for any post-processing.

\begin{figure}[t]
\centering
\includegraphics[width=0.96\linewidth]{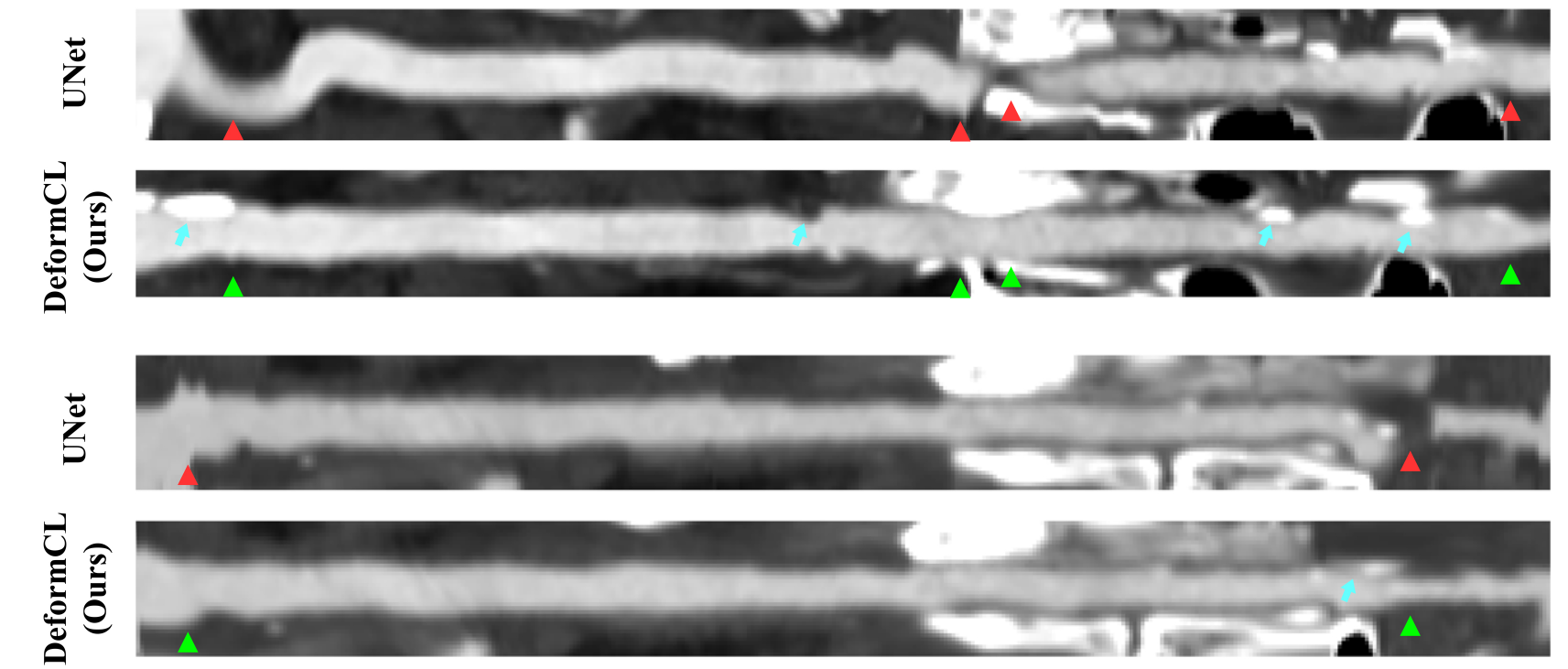}
\caption{Visual comparison of SCPR images reconstructed from the predictions of UNet and DeformCL. \textcolor{Red}{Red} triangles indicate the regions where the vessel predictions by UNet are not correct, resulting in severe errors on SCPR images, while \textcolor[rgb]{0.3,0.85,0}{green} ones indicate improved results from our DeformCL. \textcolor{Cyan}{Cyan} arrows indicate the areas with plaque and stenosis needing additional clinical diagnosis.}
\label{fig: clinical}
\end{figure}

As depicted in Figure~\ref{fig: clinical}, we compare the SCPR images reconstructed from the predictions of UNet and DeformCL. It is evident that the SCPR images generated from UNet exhibit severe bending and breaking, indicated by red arrows, which could hinder plaque diagnosis. These inaccuracies stem from the segmentation masks being fragmented by UNet in fragile regions, potentially containing plaques and stenosis. In contrast, DeformCL produces high-quality SCPR images that reveal diseased regions comprehensively and clearly. These visualizations underscore the potential of DeformCL in assisting the clinical diagnosis process.

\section{Conclusions}
In this paper, we propose DeformCL, a novel continuously defined representation for 3D vessel segmentation. To fully leverage the advantages of DeformCL, we introduce a cascaded deformation process based on an adaptive template generation strategy. Through this comprehensive training framework, DeformCL demonstrates clear superiority over previous methods. The visualizations highlight DeformCL’s potential in generating CPR images for clinical diagnosis, especially for patients with severe stenosis.

\noindent\textbf{Acknowledgement.} This work is supported by National Science and Technology Major Project (2022ZD0114902) and National Science Foundation of China (NSFC92470123, NSFC62276005).

{
    \small
    \bibliographystyle{ieeenat_fullname}
    \bibliography{main}

\begin{thebibliography}{43}
\providecommand{\natexlab}[1]{#1}
\providecommand{\url}[1]{\texttt{#1}}
\expandafter\ifx\csname urlstyle\endcsname\relax
  \providecommand{\doi}[1]{doi: #1}\else
  \providecommand{\doi}{doi: \begingroup \urlstyle{rm}\Url}\fi

\bibitem[Ara{\'u}jo et~al.(2021)Ara{\'u}jo, Cardoso, and Oliveira]{araujo2021topological}
Ricardo~J Ara{\'u}jo, Jaime~S Cardoso, and H{\'e}lder~P Oliveira.
\newblock Topological similarity index and loss function for blood vessel segmentation.
\newblock \emph{arXiv preprint arXiv:2107.14531}, 2021.

\bibitem[Bongratz et~al.(2022)Bongratz, Rickmann, P{\"o}lsterl, and Wachinger]{bongratz2022vox2cortex}
Fabian Bongratz, Anne-Marie Rickmann, Sebastian P{\"o}lsterl, and Christian Wachinger.
\newblock Vox2cortex: fast explicit reconstruction of cortical surfaces from 3d mri scans with geometric deep neural networks.
\newblock In \emph{Proceedings of the IEEE/CVF Conference on Computer Vision and Pattern Recognition}, pages 20773--20783, 2022.

\bibitem[{\c{C}}i{\c{c}}ek et~al.(2016){\c{C}}i{\c{c}}ek, Abdulkadir, Lienkamp, Brox, and Ronneberger]{cciccek20163d}
{\"O}zg{\"u}n {\c{C}}i{\c{c}}ek, Ahmed Abdulkadir, Soeren~S Lienkamp, Thomas Brox, and Olaf Ronneberger.
\newblock 3d u-net: learning dense volumetric segmentation from sparse annotation.
\newblock In \emph{Medical Image Computing and Computer-Assisted Intervention--MICCAI 2016: 19th International Conference, Athens, Greece, October 17-21, 2016, Proceedings, Part II 19}, pages 424--432. Springer, 2016.

\bibitem[Dai et~al.(2017)Dai, Qi, Xiong, Li, Zhang, Hu, and Wei]{dai2017deformable}
Jifeng Dai, Haozhi Qi, Yuwen Xiong, Yi Li, Guodong Zhang, Han Hu, and Yichen Wei.
\newblock Deformable convolutional networks.
\newblock In \emph{Proceedings of the IEEE international conference on computer vision}, pages 764--773, 2017.

\bibitem[Gharleghi et~al.(2022)Gharleghi, Adikari, Ellenberger, Ooi, Ellis, Chen, Gao, He, Hussain, Lee, Li, Ma, Nie, Oliveira, Qi, Skandarani, Vilaça, Wang, Yang, Sowmya, and Beier]{gharleghi2022automated}
Ramtin Gharleghi, Dona Adikari, Katy Ellenberger, Sze-Yuan Ooi, Chris Ellis, Chung-Ming Chen, Ruochen Gao, Yuting He, Raabid Hussain, Chia-Yen Lee, Jun Li, Jun Ma, Ziwei Nie, Bruno Oliveira, Yaolei Qi, Youssef Skandarani, João~L. Vilaça, Xiyue Wang, Sen Yang, Arcot Sowmya, and Susann Beier.
\newblock Automated segmentation of normal and diseased coronary arteries – the asoca challenge.
\newblock \emph{Computerized Medical Imaging and Graphics}, 97:\penalty0 102049, 2022.

\bibitem[Gharleghi et~al.(2023)Gharleghi, Adikari, Ellenberger, Webster, Ellis, Sowmya, Ooi, and Beier]{gharleghi2023annotated}
R Gharleghi, D Adikari, K Ellenberger, M Webster, C Ellis, A Sowmya, S Ooi, and S Beier.
\newblock Annotated computed tomography coronary angiogram images and associated data of normal and diseased arteries.
\newblock \emph{Scientific Data}, 10\penalty0 (1):\penalty0 128, 2023.

\bibitem[Graham and Hell(1985)]{graham1985history}
Ronald~L Graham and Pavol Hell.
\newblock On the history of the minimum spanning tree problem.
\newblock \emph{Annals of the History of Computing}, 7\penalty0 (1):\penalty0 43--57, 1985.

\bibitem[Guimaraes et~al.(2016)Guimaraes, Wigdahl, and Ruggeri]{guimaraes2016fast}
Pedro Guimaraes, Jeffrey Wigdahl, and Alfredo Ruggeri.
\newblock A fast and efficient technique for the automatic tracing of corneal nerves in confocal microscopy.
\newblock \emph{Translational vision science \& technology}, 5\penalty0 (5), 2016.

\bibitem[Hu et~al.(2019)Hu, Li, Samaras, and Chen]{hu2019topology}
Xiaoling Hu, Fuxin Li, Dimitris Samaras, and Chao Chen.
\newblock Topology-preserving deep image segmentation.
\newblock \emph{Advances in neural information processing systems}, 32, 2019.

\bibitem[Izzo et~al.(2018)Izzo, Steinman, Manini, and Antiga]{izzo2018vascular}
Richard Izzo, David Steinman, Simone Manini, and Luca Antiga.
\newblock The vascular modeling toolkit: a python library for the analysis of tubular structures in medical images.
\newblock \emph{Journal of Open Source Software}, 3\penalty0 (25):\penalty0 745, 2018.

\bibitem[Kanitsar et~al.(2002)Kanitsar, Fleischmann, Wegenkittl, Felkel, and Groller]{kanitsar2002cpr}
Armin Kanitsar, Dominik Fleischmann, Rainer Wegenkittl, Petr Felkel, and Eduard Groller.
\newblock \emph{CPR-curved planar reformation}.
\newblock IEEE, 2002.

\bibitem[Kipf and Welling(2016)]{kipf2016semi}
Thomas~N Kipf and Max Welling.
\newblock Semi-supervised classification with graph convolutional networks.
\newblock \emph{arXiv preprint arXiv:1609.02907}, 2016.

\bibitem[Kong et~al.(2020)Kong, Wang, Bai, Lu, Gao, Cao, Xia, Song, and Yin]{kong2020learning}
Bin Kong, Xin Wang, Junjie Bai, Yi Lu, Feng Gao, Kunlin Cao, Jun Xia, Qi Song, and Youbing Yin.
\newblock Learning tree-structured representation for 3d coronary artery segmentation.
\newblock \emph{Computerized Medical Imaging and Graphics}, 80:\penalty0 101688, 2020.

\bibitem[Kong et~al.(2021)Kong, Wilson, and Shadden]{kong2021deep}
Fanwei Kong, Nathan Wilson, and Shawn Shadden.
\newblock A deep-learning approach for direct whole-heart mesh reconstruction.
\newblock \emph{Medical image analysis}, 74:\penalty0 102222, 2021.

\bibitem[Lee et~al.(1994)Lee, Kashyap, and Chu]{lee1994building}
Ta-Chih Lee, Rangasami~L Kashyap, and Chong-Nam Chu.
\newblock Building skeleton models via 3-d medial surface axis thinning algorithms.
\newblock \emph{CVGIP: Graphical Models and Image Processing}, 56\penalty0 (6):\penalty0 462--478, 1994.

\bibitem[Leipsic et~al.(2014)Leipsic, Abbara, Achenbach, Cury, Earls, Mancini, Nieman, Pontone, and Raff]{leipsic2014scct}
Jonathon Leipsic, Suhny Abbara, Stephan Achenbach, Ricardo Cury, James~P Earls, GB~John Mancini, Koen Nieman, Gianluca Pontone, and Gilbert~L Raff.
\newblock Scct guidelines for the interpretation and reporting of coronary ct angiography: a report of the society of cardiovascular computed tomography guidelines committee.
\newblock \emph{Journal of cardiovascular computed tomography}, 8\penalty0 (5):\penalty0 342--358, 2014.

\bibitem[Long et~al.(2015)Long, Shelhamer, and Darrell]{long2015fully}
Jonathan Long, Evan Shelhamer, and Trevor Darrell.
\newblock Fully convolutional networks for semantic segmentation.
\newblock In \emph{Proceedings of the IEEE conference on computer vision and pattern recognition}, pages 3431--3440, 2015.

\bibitem[Loshchilov and Hutter(2017)]{loshchilov2017decoupled}
Ilya Loshchilov and Frank Hutter.
\newblock Decoupled weight decay regularization.
\newblock \emph{arXiv preprint arXiv:1711.05101}, 2017.

\bibitem[Milletari et~al.(2016)Milletari, Navab, and Ahmadi]{milletari2016v}
Fausto Milletari, Nassir Navab, and Seyed-Ahmad Ahmadi.
\newblock V-net: Fully convolutional neural networks for volumetric medical image segmentation.
\newblock In \emph{2016 fourth international conference on 3D vision (3DV)}, pages 565--571. Ieee, 2016.

\bibitem[Paszke et~al.(2019)Paszke, Gross, Massa, Lerer, Bradbury, Chanan, Killeen, and Zeming~Lin]{adam19torch}
Adam Paszke, Sam Gross, Francisco Massa, Adam Lerer, James Bradbury, Gregory Chanan, Trevor Killeen, and et~al. Zeming~Lin.
\newblock Pytorch: An imperative style, high-performance deep learning library.
\newblock In \emph{NeurIPS}, 2019.

\bibitem[Piccinelli et~al.(2009)Piccinelli, Veneziani, Steinman, Remuzzi, and Antiga]{piccinelli2009framework}
Marina Piccinelli, Alessandro Veneziani, David~A Steinman, Andrea Remuzzi, and Luca Antiga.
\newblock A framework for geometric analysis of vascular structures: application to cerebral aneurysms.
\newblock \emph{IEEE transactions on medical imaging}, 28\penalty0 (8):\penalty0 1141--1155, 2009.

\bibitem[Podobnik et~al.(2023)Podobnik, Strojan, Peterlin, Ibragimov, and Vrtovec]{podobnik2023han}
Ga{\v{s}}per Podobnik, Primo{\v{z}} Strojan, Primo{\v{z}} Peterlin, Bulat Ibragimov, and Toma{\v{z}} Vrtovec.
\newblock Han-seg: The head and neck organ-at-risk ct and mr segmentation dataset.
\newblock \emph{Medical physics}, 50\penalty0 (3):\penalty0 1917--1927, 2023.

\bibitem[Qi et~al.(2023)Qi, He, Qi, Zhang, and Yang]{qi2023dynamic}
Yaolei Qi, Yuting He, Xiaoming Qi, Yuan Zhang, and Guanyu Yang.
\newblock Dynamic snake convolution based on topological geometric constraints for tubular structure segmentation.
\newblock In \emph{Proceedings of the IEEE/CVF International Conference on Computer Vision}, pages 6070--6079, 2023.

\bibitem[Roger et~al.(2011)Roger, Go, Lloyd-Jones, Adams, Berry, Brown, Carnethon, Dai, De~Simone, Ford, et~al.]{roger2011heart}
V{\'e}ronique~L Roger, Alan~S Go, Donald~M Lloyd-Jones, Robert~J Adams, Jarett~D Berry, Todd~M Brown, Mercedes~R Carnethon, Shifan Dai, Giovanni De~Simone, Earl~S Ford, et~al.
\newblock Heart disease and stroke statistics—2011 update: a report from the american heart association.
\newblock \emph{Circulation}, 123\penalty0 (4):\penalty0 e18--e209, 2011.

\bibitem[Ronneberger et~al.(2015)Ronneberger, Fischer, and Brox]{ronneberger2015u}
Olaf Ronneberger, Philipp Fischer, and Thomas Brox.
\newblock U-net: Convolutional networks for biomedical image segmentation.
\newblock In \emph{Medical Image Computing and Computer-Assisted Intervention--MICCAI 2015: 18th International Conference, Munich, Germany, October 5-9, 2015, Proceedings, Part III 18}, pages 234--241. Springer, 2015.

\bibitem[Schaap et~al.(2009)Schaap, Metz, van Walsum, van~der Giessen, Weustink, Mollet, Bauer, Bogunovi{\'c}, Castro, Deng, et~al.]{schaap2009standardized}
Michiel Schaap, Coert~T Metz, Theo van Walsum, Alina~G van~der Giessen, Annick~C Weustink, Nico~R Mollet, Christian Bauer, Hrvoje Bogunovi{\'c}, Carlos Castro, Xiang Deng, et~al.
\newblock Standardized evaluation methodology and reference database for evaluating coronary artery centerline extraction algorithms.
\newblock \emph{Medical image analysis}, 13\penalty0 (5):\penalty0 701--714, 2009.

\bibitem[Shin et~al.(2019)Shin, Lee, Yun, and Lee]{shin2019deep}
Seung~Yeon Shin, Soochahn Lee, Il~Dong Yun, and Kyoung~Mu Lee.
\newblock Deep vessel segmentation by learning graphical connectivity.
\newblock \emph{Medical image analysis}, 58:\penalty0 101556, 2019.

\bibitem[Shit et~al.(2021)Shit, Paetzold, Sekuboyina, Ezhov, Unger, Zhylka, Pluim, Bauer, and Menze]{shit2021cldice}
Suprosanna Shit, Johannes~C Paetzold, Anjany Sekuboyina, Ivan Ezhov, Alexander Unger, Andrey Zhylka, Josien~PW Pluim, Ulrich Bauer, and Bjoern~H Menze.
\newblock cldice-a novel topology-preserving loss function for tubular structure segmentation.
\newblock In \emph{Proceedings of the IEEE/CVF Conference on Computer Vision and Pattern Recognition}, pages 16560--16569, 2021.

\bibitem[Taylor and Steinman(2010)]{taylor2010image}
Charles~A Taylor and David~A Steinman.
\newblock Image-based modeling of blood flow and vessel wall dynamics: applications, methods and future directions: Sixth international bio-fluid mechanics symposium and workshop, march 28--30, 2008 pasadena, california.
\newblock \emph{Annals of biomedical engineering}, 38:\penalty0 1188--1203, 2010.

\bibitem[Vaswani et~al.(2017)Vaswani, Shazeer, Parmar, Uszkoreit, Jones, Gomez, Kaiser, and Polosukhin]{vaswani2017attention}
Ashish Vaswani, Noam Shazeer, Niki Parmar, Jakob Uszkoreit, Llion Jones, Aidan~N Gomez, {\L}ukasz Kaiser, and Illia Polosukhin.
\newblock Attention is all you need.
\newblock \emph{Advances in neural information processing systems}, 30, 2017.

\bibitem[Wang et~al.(2022)Wang, Zhang, Zhao, Liu, Chen, and Wang]{wang2022pointscatter}
Dong Wang, Zhao Zhang, Ziwei Zhao, Yuhang Liu, Yihong Chen, and Liwei Wang.
\newblock Pointscatter: Point set representation for tubular structure extraction.
\newblock In \emph{European Conference on Computer Vision}, pages 366--383. Springer, 2022.

\bibitem[Wang et~al.(2018)Wang, Zhang, Li, Fu, Liu, and Jiang]{wang2018pixel2mesh}
Nanyang Wang, Yinda Zhang, Zhuwen Li, Yanwei Fu, Wei Liu, and Yu-Gang Jiang.
\newblock Pixel2mesh: Generating 3d mesh models from single rgb images.
\newblock In \emph{Proceedings of the European conference on computer vision (ECCV)}, pages 52--67, 2018.

\bibitem[Wang et~al.(2020)Wang, Wei, Liu, Chen, Zhou, Shen, Fishman, and Yuille]{wang2020deep}
Yan Wang, Xu Wei, Fengze Liu, Jieneng Chen, Yuyin Zhou, Wei Shen, Elliot~K Fishman, and Alan~L Yuille.
\newblock Deep distance transform for tubular structure segmentation in ct scans.
\newblock In \emph{Proceedings of the IEEE/CVF Conference on Computer Vision and Pattern Recognition}, pages 3833--3842, 2020.

\bibitem[Wickramasinghe et~al.(2020)Wickramasinghe, Remelli, Knott, and Fua]{wickramasinghe2020voxel2mesh}
Udaranga Wickramasinghe, Edoardo Remelli, Graham Knott, and Pascal Fua.
\newblock Voxel2mesh: 3d mesh model generation from volumetric data.
\newblock In \emph{Medical Image Computing and Computer Assisted Intervention--MICCAI 2020: 23rd International Conference, Lima, Peru, October 4--8, 2020, Proceedings, Part IV 23}, pages 299--308. Springer, 2020.

\bibitem[Wu et~al.(2024)Wu, Chen, Liu, Yue, and Zhuang]{wu2024deep}
Qian Wu, Yufei Chen, Wei Liu, Xiaodong Yue, and Xiahai Zhuang.
\newblock Deep closing: Enhancing topological connectivity in medical tubular segmentation.
\newblock \emph{IEEE Transactions on Medical Imaging}, 2024.

\bibitem[Yang et~al.(2022)Yang, Wickramasinghe, Ni, and Fua]{yang2022implicitatlas}
Jiancheng Yang, Udaranga Wickramasinghe, Bingbing Ni, and Pascal Fua.
\newblock Implicitatlas: learning deformable shape templates in medical imaging.
\newblock In \emph{Proceedings of the IEEE/CVF Conference on Computer Vision and Pattern Recognition}, pages 15861--15871, 2022.

\bibitem[Zeng et~al.(2023)Zeng, Wu, Lin, Xie, Hong, Huang, Zhuang, Bi, Pan, Ullah, et~al.]{zeng2023imagecas}
An Zeng, Chunbiao Wu, Guisen Lin, Wen Xie, Jin Hong, Meiping Huang, Jian Zhuang, Shanshan Bi, Dan Pan, Najeeb Ullah, et~al.
\newblock Imagecas: A large-scale dataset and benchmark for coronary artery segmentation based on computed tomography angiography images.
\newblock \emph{Computerized Medical Imaging and Graphics}, 109:\penalty0 102287, 2023.

\bibitem[Zhang et~al.(2022)Zhang, Zhang, Ma, Xue, Hu, Wu, Zhan, Feng, and Shen]{zhang2022progressive}
Xiao Zhang, Jingyang Zhang, Lei Ma, Peng Xue, Yan Hu, Dijia Wu, Yiqiang Zhan, Jun Feng, and Dinggang Shen.
\newblock Progressive deep segmentation of coronary artery via hierarchical topology learning.
\newblock In \emph{International Conference on Medical Image Computing and Computer-Assisted Intervention}, pages 391--400. Springer, 2022.

\bibitem[Zhang et~al.(2023)Zhang, Zhao, Wang, Zhao, Liu, Liu, and Wang]{zhang2023topology}
Zhixing Zhang, Ziwei Zhao, Dong Wang, Shishuang Zhao, Yuhang Liu, Jia Liu, and Liwei Wang.
\newblock Topology-preserving automatic labeling of coronary arteries via anatomy-aware connection classifier.
\newblock In \emph{International Conference on Medical Image Computing and Computer-Assisted Intervention}, pages 759--769. Springer, 2023.

\bibitem[Zhang et~al.(2024)Zhang, Zhao, Wang, and Wang]{zhang2024graphmorph}
Zhao Zhang, Ziwei Zhao, Dong Wang, and Liwei Wang.
\newblock Graphmorph: Tubular structure extraction by morphing predicted graphs.
\newblock \emph{Advances in Neural Information Processing Systems}, 37:\penalty0 68472--68499, 2024.

\bibitem[Zhao et~al.(2022)Zhao, Liang, Pan, Zhang, Wu, Hu, and Yu]{zhao2022graph}
Gangming Zhao, Kongming Liang, Chengwei Pan, Fandong Zhang, Xianpeng Wu, Xinyang Hu, and Yizhou Yu.
\newblock Graph convolution based cross-network multiscale feature fusion for deep vessel segmentation.
\newblock \emph{IEEE Transactions on Medical Imaging}, 42\penalty0 (1):\penalty0 183--195, 2022.

\bibitem[Zhao et~al.(2021)Zhao, Cao, Yao, Nogues, Lu, Huang, Xiao, Yin, and Zhang]{zhao20213d}
Tianyi Zhao, Kai Cao, Jiawen Yao, Isabella Nogues, Le Lu, Lingyun Huang, Jing Xiao, Zhaozheng Yin, and Ling Zhang.
\newblock 3d graph anatomy geometry-integrated network for pancreatic mass segmentation, diagnosis, and quantitative patient management.
\newblock In \emph{Proceedings of the IEEE/CVF conference on computer vision and pattern recognition}, pages 13743--13752, 2021.

\bibitem[Zhu et~al.(2020)Zhu, Su, Lu, Li, Wang, and Dai]{zhu2020deformable}
Xizhou Zhu, Weijie Su, Lewei Lu, Bin Li, Xiaogang Wang, and Jifeng Dai.
\newblock Deformable detr: Deformable transformers for end-to-end object detection.
\newblock \emph{arXiv preprint arXiv:2010.04159}, 2020.

\end{thebibliography}
}

\clearpage

\renewcommand{\thesection}{\Alph{section}} 
\setcounter{section}{0}

We will elaborate on the details about datasets, implementation and more experimental analysis in supplementary materials.
\section{Dataset Details}
We use four datasets to evaluate our proposed framework. For the \textbf{HaN-Seg}~\cite{podobnik2023han} dataset, we randomly split the publicly available 42 CT scans into 31 and 11 images as train and test data, and the MR images were not used. Left and right carotid arteries are utilized as segmentation targets. In the case of the \textbf{HNCTA} dataset containing 358 head and neck CT scans, clinical experts provide accurate segmentation labels for nine categories of vessels, including the left and right vertebral arteries (VA), internal carotid arteries (ICA), common carotid arteries (CCA), posterior communicating arteries (PoCA), and basilar artery (BA). The dataset is randomly split into 242, 58, and 58 images as train, validation, and test data.
For the  \textbf{ASOCA}~\cite{gharleghi2022automated, gharleghi2023annotated} dataset, we split the dataset randomly into 33 training and 7 testing images. For the \textbf{ImageCAS}~\cite{zeng2023imagecas} dataset, we fulfill the vessel category annotations for 350 cases with experienced radiologists and split it into 235, 50, and 65 for train, validation, and test data. We evaluate the proposed method on the three main coronary arteries (LAD, LCX, and RCA) for all CCTA datasets including ASOCA and ImageCAS.

\section{Additional Implementation Details}

During the training process, we employ RandomCrop and RandomFlip as data augmentations to mitigate overfitting, with consistent settings applied across all methods. For larger datasets such as ImageCAS~\cite{zeng2023imagecas} and HNCTA, we train our model and all baselines for 12,000 iterations on 8 NVIDIA GeForce RTX 3090 GPUs. For smaller datasets like ASOCA~\cite{gharleghi2022automated, gharleghi2023annotated} and HaN-Seg~\cite{podobnik2023han}, we train our model and all baselines for 4,000 iterations. The batch size is set to 8 by default for all experiments. We utilize the AdamW Optimizer~\cite{loshchilov2017decoupled} with a base learning rate of $1\times10^{-3}$. Additionally, we employ a cosine learning rate schedule with warmup. All models are implemented based on PyTorch~\cite{adam19torch}.

We provide the required resources and runtime analusis on 3D UNet in Table~\ref{tab:resource}, and we can conclude that the increase of DeformCL in resource usage is minimal.

\begin{table}[h]
    \caption{Computational costs and runtime analysis.}
    \centering
    \scalebox{0.57}{
    \begin{tabular}{c|c|c|c|c}
    \toprule
     Method & Params (M) & FLOPs (G) & GPU Memory (MB) & Inference Time (s/img)\\
    \midrule
    SoftDice & 2.08 & 3.11 &3046 &0.12 \\
    \midrule
    Ours(SoftDice) & 2.20 & 4.03 & 3228&0.18\\
    \bottomrule
    \end{tabular}
    }
    \label{tab:resource}
\end{table}

\section{Analysis of Centerline Deformation Process}
\label{cline deform}
\begin{figure}[t]
\centering
\includegraphics[width=0.95\linewidth]{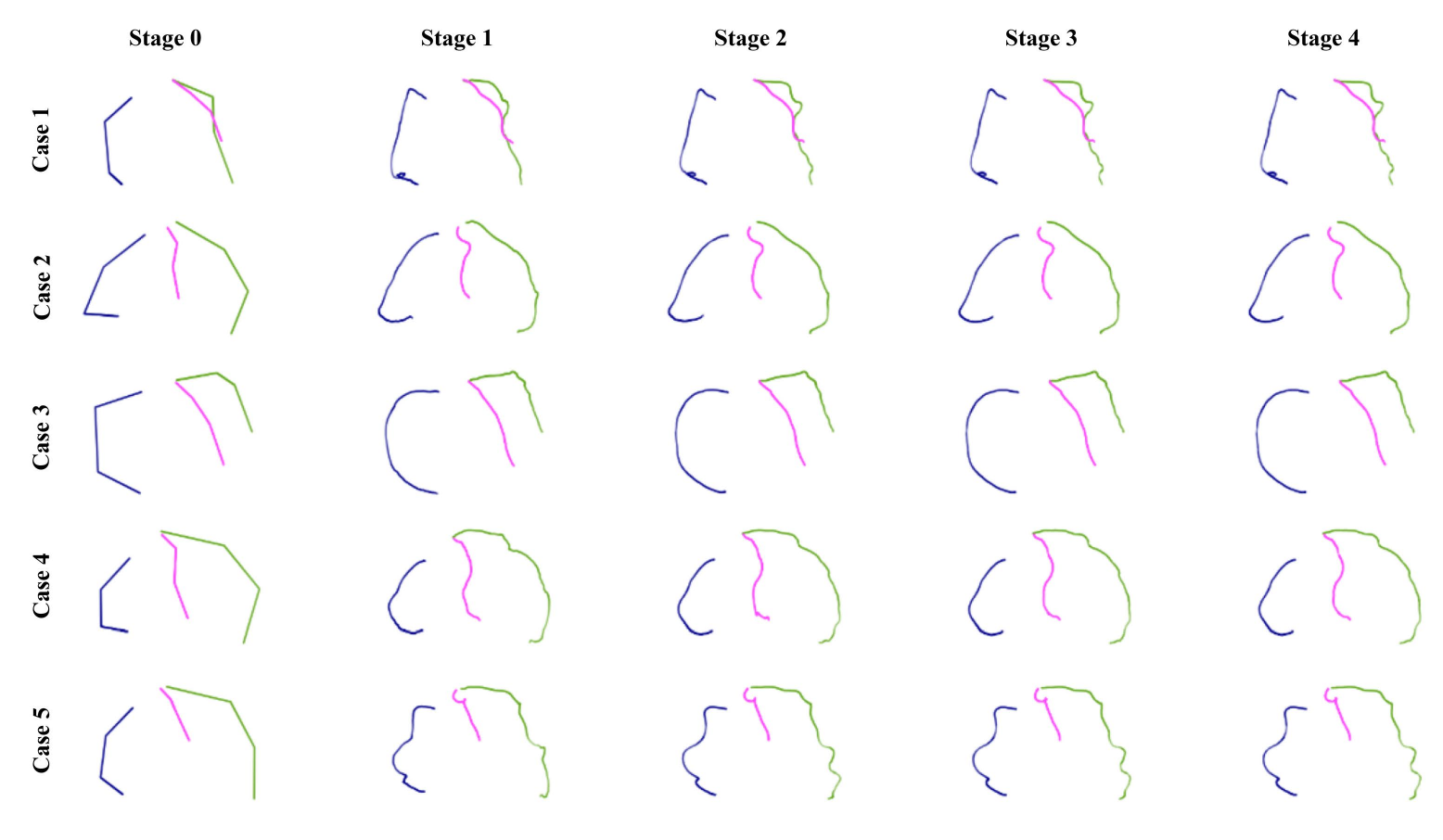}
\caption{
Visualizations of the centerline deformation process on the ImageCAS dataset. Each row corresponds to a different case, and each column represents a different deformation stage. The first subfigure in each row (Stage 0) displays the centerline template generated using linear interpolation, while subsequent subfigures show the centerline predictions after deformation stages 1, 2, 3, and 4, respectively.}
\label{fig 3}
\end{figure}

We present visualizations of the centerline deformation process on the ImageCAS dataset~\cite{zeng2023imagecas}, aimed at enhancing understanding of DeformCL, as depicted in Figure~\ref{fig 3}. Stage 0 represents the centerline templates generated by linear interpolation on control points. It is observed that the core structure of the centerlines typically stabilizes within the initial one or two stages, with subsequent stages primarily focused on refining finer details. Furthermore, we note that the predicted centerlines naturally possess sufficient smoothness, which proves beneficial for the subsequent clinical tasks such as CPR~\cite{kanitsar2002cpr} image generation.

\section{Additional Experimental Results}

 For broader comparisons, we include Transformer backbones Swin UNETR in Table~\ref{tab:seg_swin}. We also provide more ablation studies as follows.
 
\begin{table}[t]
\centering
\caption{Results of segmentation and centerline on HaN-Seg dataset with Swin UNETR as backbone.}
\label{tab:seg_swin}
\renewcommand\arraystretch{1.2}
\scalebox{0.45}{
\begin{tabular}{c | l | c c  || c c | c c c| c }
\toprule
\mrtwo{Backbone}	&\mrtwo{Method}	& \multicolumn{2}{c||}{Centerline Scores} & \multicolumn{2}{c|}{Volumetric Scores (\%) $\uparrow$}& \multicolumn{3}{c|}{Topological Error $\downarrow$} & Distance $\downarrow$ \\
\cline{3-10}
&&	F1 (\%) $\uparrow$ & Chamfer $\downarrow$ &Dice	&clDice	&$\beta_0$	&$\beta_1$	& Euler & HD95	\\
\midrule										
    {Swin UNETR}&softDice &91.72&11.37& 77.05 & 89.06  & 1.815 & 0.206 & 1.738 & 4.624 \\
    \rowcolor{mygray}\cellcolor{white}{Swin UNETR}&\textbf{Ours} (softDice) &94.54&7.53& 79.45 & 89.46  & 1.410 & 0.051 & 1.388 & 3.701 \\
    {Swin UNETR}& clDice	&92.79&15.67& 78.41 & 89.50  & 2.116 & 0.167 & 1.977 & 4.513 \\
 \rowcolor{mygray}\cellcolor{white}{Swin UNETR}&\textbf{Ours} (clDice) &94.11&7.10& 79.29 & 89.62 & 1.982 & 0.096& 1.905 & 3.718 \\
 
\bottomrule										
\end{tabular}

}
\end{table}

\begin{figure}[bp]
\centering
\includegraphics[width=0.85\linewidth]{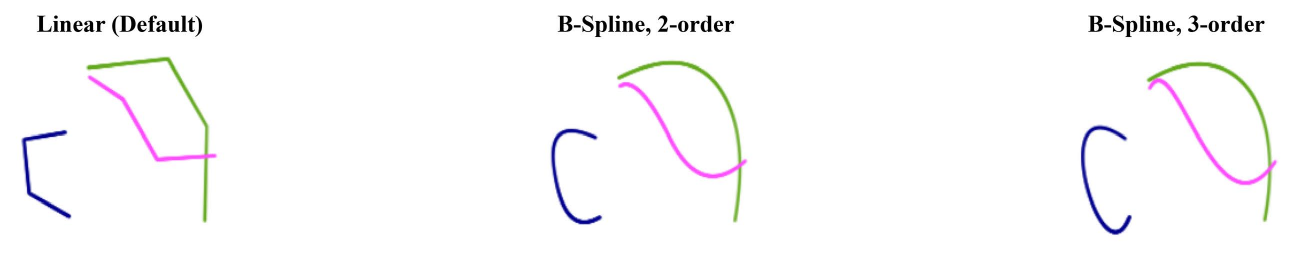}
\caption{
Visualization of centerline templates obtained through different interpolation methods: linear interpolation, second-order B-Spline, third-order B-Spline.
}
\label{fig 4}
\end{figure}

\noindent\textbf{Curvilinear Interpolation Method.} 
During adaptive template generation, curvilinear interpolation is essential for generating the initial centerline template based on control points. 
In this subsection, we investigate the efficacy of different interpolation methods, encompassing second- and third-order B-Spline Interpolation, as well as Linear Interpolation, which serves as our default setting owing to its simplicity. Figure \ref{fig 4} provides an illustration of the aforementioned interpolation methods. Experimental results in Table~\ref{tab 2} demonstrate that different interpolation methods yield similar performance in vessel segmentation. This suggests that the overall framework is robust to initial templates, and several deformation stages are adequate to deform the centerline templates towards the ground truth centerlines effectively. This observation aligns with the analysis in Section \ref{cline deform} -- after the initial one or two stages, the core structure of centerlines will be stabilized.

\begin{table}[t]
\centering
\caption{Ablation on curvilinear interpolation method on HaN-Seg dataset.} 
\label{tab 2}
\renewcommand\arraystretch{1.2}
\scalebox{0.89}{
\begin{tabular}{c|cc}
\toprule
\multirow{2}{*}{Interpolation Methods}                 &  \multicolumn{2}{c}{Volumetric Scores (\%) $\uparrow$}               \\ \cline{2-3}                   
            & \multicolumn{1}{c}{Dice} & \multicolumn{1}{c}{clDice} \\ \hline

B-Spline, 2-order                                                              & \textbf{79.95}                 & 88.20                  \\
B-Spline, 3-order                                                              & 79.53                & 88.10               \\
\rowcolor{mygray} Linear (Default)               & 79.58     & \textbf{88.86}            \\ \bottomrule

\end{tabular}
}
\end{table}

\noindent\textbf{Interaction Approach.} As discussed in the main paper, previous methods~\cite{qi2023dynamic, shit2021cldice, wang2020deep, wang2022pointscatter} lack an effective approach for curvilinear feature aggregation due to their discrete representations. In contrast, DeformCL  inherently exhibits a graph structure, enabling point feature interaction along the tubular curve efficiently. Table~\ref{tab: ablation inter} compares different interaction approaches, including Graph Convolutional Network (GCN)~\cite{kipf2016semi} and Transformer~\cite{vaswani2017attention}. The results indicate that the improvement with Transformer is more significant than with GCN, possibly due to its powerful long-distance modeling capability.

\begin{table}[t]
\centering
\caption{Ablation on interaction approach on HaN-Seg dataset.} 
\label{tab: ablation inter}
\renewcommand\arraystretch{1.2}
\scalebox{0.89}{
\begin{tabular}{c|cc}
\toprule
\multirow{2}{*}{Interaction Approaches}                 &  \multicolumn{2}{c}{Volumetric Scores (\%) $\uparrow$}               \\ \cline{2-3}                   
            & \multicolumn{1}{c}{Dice} & \multicolumn{1}{c}{clDice} \\ \hline

$\emptyset$                                                             & 78.75                   & 88.40                      \\
GCN                                                              & 78.77                   & 88.61                    \\
\rowcolor{mygray} Transformer (Default)               & \textbf{79.58}     & \textbf{88.86}            \\ \bottomrule

\end{tabular}
}
\end{table}

\section{Additional Qualitative Results}

\begin{figure}[t]
\centering
\includegraphics[width=0.9\linewidth]{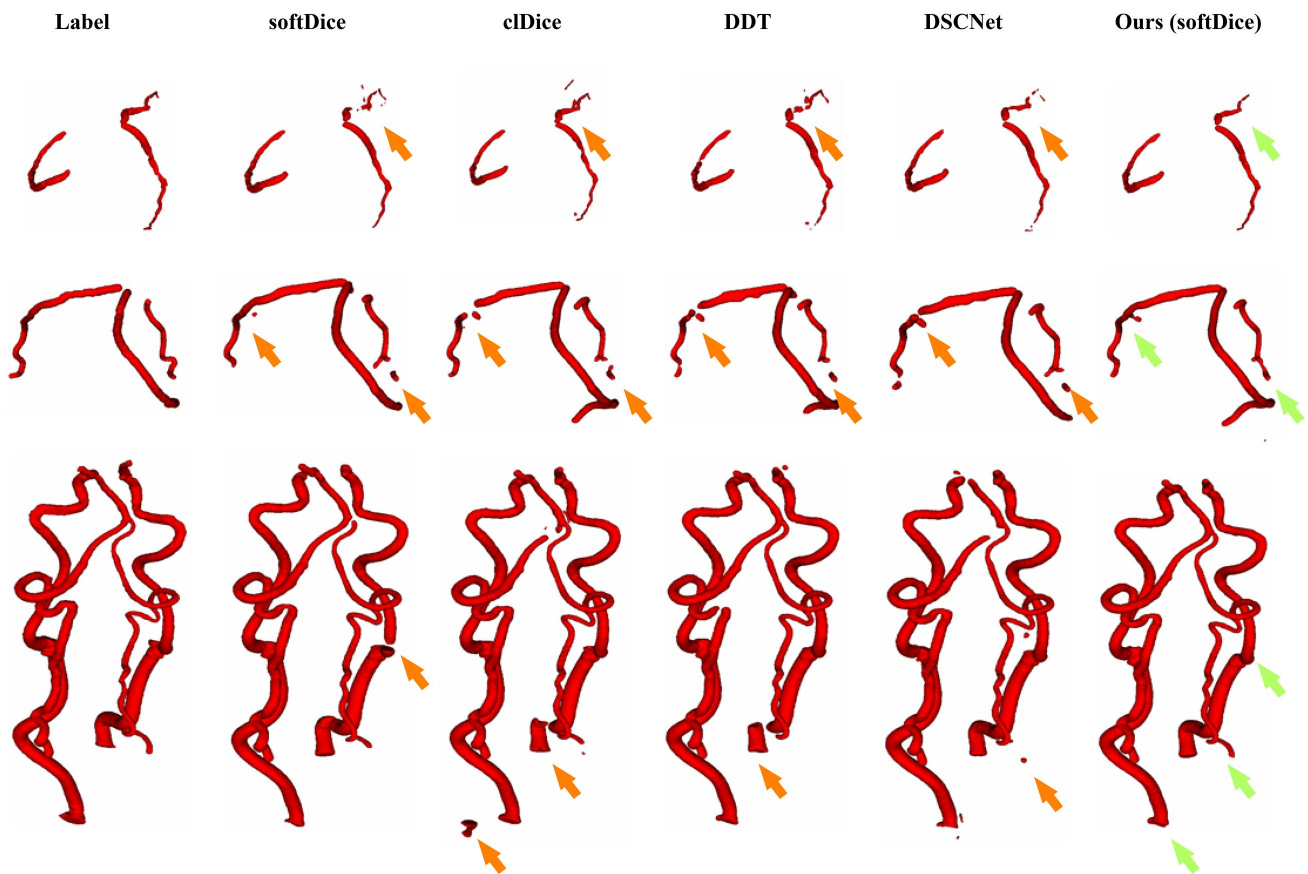}
\caption{Additional visualizations of the predictions from different methods on 3D medical
image datasets. The \textcolor{Orange}{orange} arrows point out the areas where previous methods exhibit errors, while the \textcolor[rgb]{0,0.9,0}{green} arrows point out the improved results from our approach.}
\label{fig 1}
\end{figure}

\begin{figure}[t]
\centering
\includegraphics[width=0.8\linewidth]{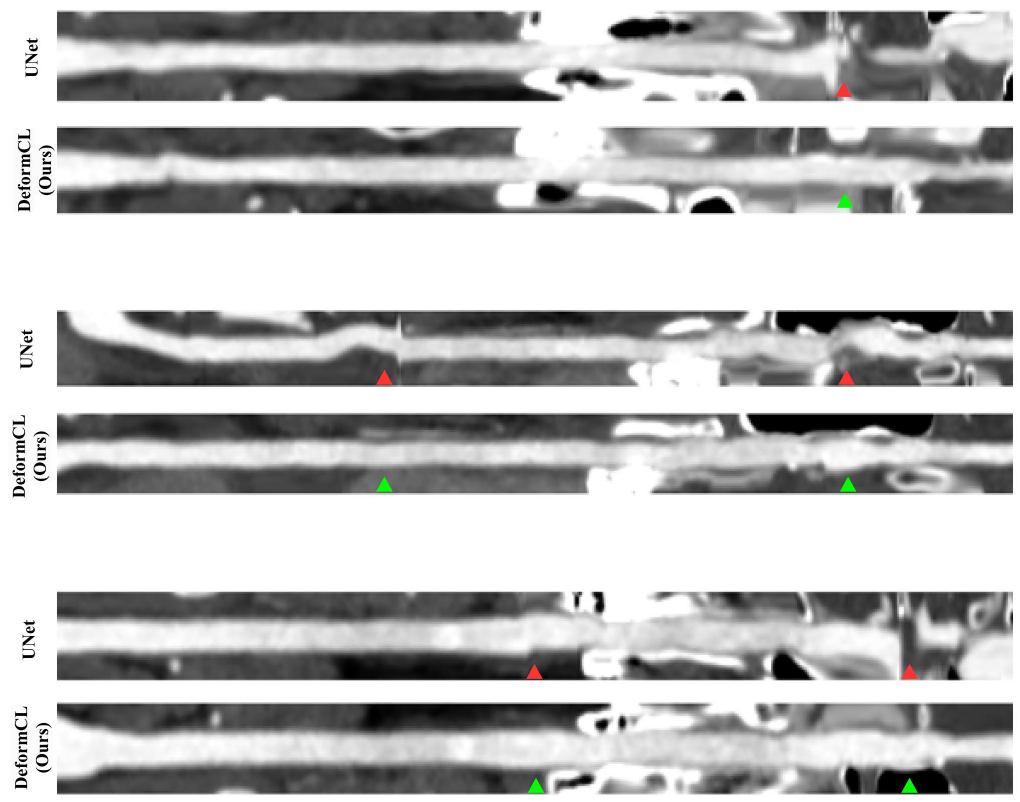}
\caption{Additional visual comparison of SCPR images reconstructed from the predictions of UNet and DeformCL. \textcolor{Red}{Red} triangles indicate the regions where the vessel predictions by UNet are not correct, resulting in severe errors on SCPR images, while \textcolor[rgb]{0,0.9,0}{green} ones indicate the improved results from our DeformCL.}
\label{fig 2}
\end{figure}

In this section, we provide additional qualitative results including the visual comparison of segmentation results and SCPR images in Figure~\ref{fig 1} and Figure~\ref{fig 2}.

\begin{figure}[t]
\centering
\includegraphics[width=1.0\linewidth]{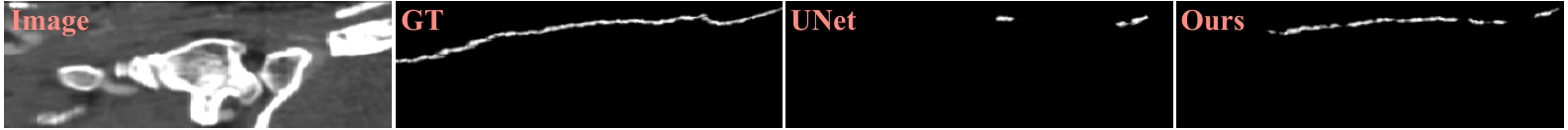}
\caption{Performance of centerlines on occluded vessels.}
\label{fig: occluded}
\end{figure}

Moreover, we present a failure case involving almost completely occluded vessels, which are highly challenging for deep learning models due to their near invisibility. As shown in Figure~\ref{fig: occluded}, our method produces fragmented predictions. However, it outperforms the previous mask-based representation, thanks to its continuous property.

\end{document}